\documentclass[11pt, a4paper, onecolumn, logo, copyright]{hunyuan_report}

\usepackage[authoryear, sort&compress, round]{natbib}
\usepackage{amsmath}
\usepackage{amssymb}
\usepackage{multirow}
\usepackage{array}
\usepackage{enumitem}
\usepackage{booktabs}
\usepackage{graphicx}
\usepackage{wrapfig}
\usepackage{placeins}
\usepackage{microtype}
\usepackage{listings}
\usepackage{colortbl}
\usepackage[listings, skins, breakable]{tcolorbox}

\graphicspath{{images/}}
\definecolor{oursbg}{RGB}{228,239,252}
\definecolor{groupbg}{RGB}{242,242,242}
\definecolor{changedbg}{RGB}{255,240,219}

\definecolor{promptframe}{RGB}{70,100,140}
\definecolor{promptback}{RGB}{245,248,252}
\newtcblisting{promptbox}[1]{
  enhanced, breakable, listing only,
  colback=promptback, colframe=promptframe, coltitle=white,
  title={#1}, fonttitle=\bfseries\small,
  boxrule=0.6pt, arc=2pt, outer arc=2pt,
  left=3pt, right=3pt, top=2pt, bottom=2pt,
  before skip=4pt, after skip=5pt,
  listing options={basicstyle=\ttfamily\scriptsize, breaklines=true, breakatwhitespace=true, columns=fullflexible, keepspaces=true, showstringspaces=false},
}

\newcommand{\std}[1]{\textsubscript{$\pm$#1}}
\newcommand{\lfone}{LongForge-1}
\newcommand{\lfn}{LongForge-N}

\newcommand{\lficon}{\raisebox{-0.26\height}{\includegraphics[height=1.35em]{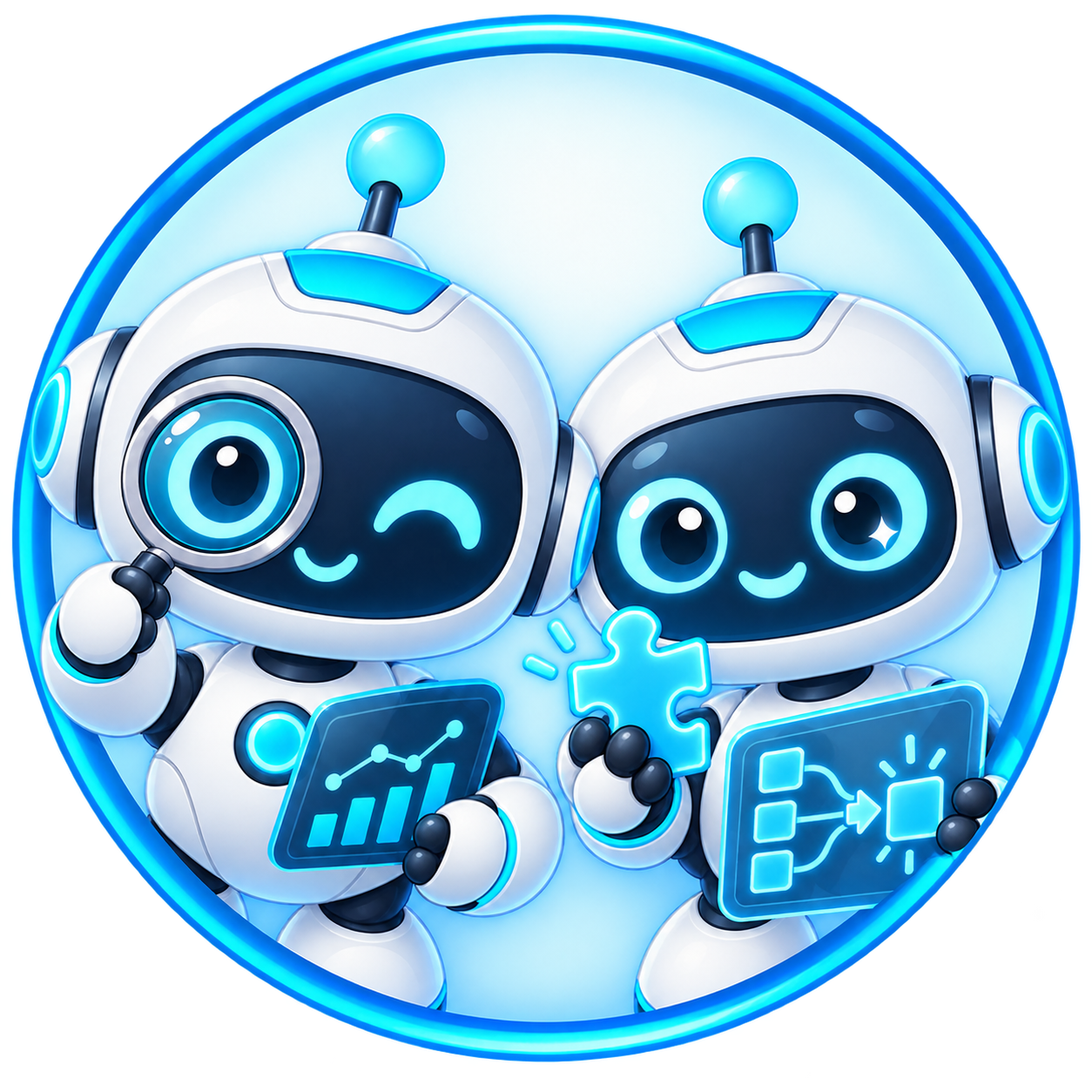}}}
\newcommand{\lfname}{\textcolor{gblue9}{\fontfamily{pag}\selectfont LongForge}}
\title{\texorpdfstring{\lficon\hspace{0.3em}\lfname}{LongForge}: Measuring and Improving Long-Context \\
Instruction Data via Evidence--Assertion Reasoning Graphs}
\runningtitle{LongForge: Measuring and Improving Long-Context Instruction Data}

\author{
  Ziyang Chen\textsuperscript{*},
  Junlong Jia\textsuperscript{*},
  Xing Wu\textsuperscript{\textdagger},
  Tinghao Yu\textsuperscript{\textdagger},
  and Songlin Hu\\
  \vspace{0.3cm}
  \normalsize
  Hunyuan Team, Tencent \quad Chinese Academy of Sciences
}

\begin{abstract}
Existing quality control for long-context instruction-tuning data is a one-shot judgment inside the generation pipeline that leaves only a scalar, revealing neither which document parts an answer uses nor how many reasoning steps it writes down; the data can thus be neither audited nor controlled.
We formalize data quality as a property of the alignment structure among document, question, and answer, and build LongForge, which pairs an analyzer that measures long-context data with a synthesizer that generates it. The analyzer reverse-reconstructs each sample into an evidence--assertion reasoning graph and programmatically computes four dimensions on it: breadth, depth, correctness, and completeness. Every reading traces back to specific evidence spans and reasoning steps, making data auditable.
Measuring seven mainstream datasets reveals two previously unquantified defects: breadth barely grows with document length, and depth is generally low, the two being nearly orthogonal.
The synthesizer then embeds the analyzer into generation, gating on correctness and completeness and deferring breadth and depth to a training-time mixture over measured values, which makes the training distribution controllable.
On two backbones, its two mixtures, \lfone{} and \lfn{}, outperform the strongest baseline by 2.2--4.7 points on average over six long-context benchmarks, and in a controlled comparison that holds the gate and the candidate pool fixed, raising breadth and depth alone is worth a further 1.16 points, making the structural distribution of training data a lever on downstream quality rather than only a description of it.
\end{abstract}

\begin{document}

\maketitle
\begingroup
\renewcommand{\thefootnote}{*}
\footnotetext{Equal contribution. \ \textsuperscript{\textdagger}Corresponding author. Correspondence to \texttt{\{ucaswu,maxwellyu\}@tencent.com}.}
\endgroup
\setcounter{footnote}{0}

% Teaser on page 1, directly under the abstract box. Not a float, so it stays
% put -- but a bare center block can be broken between the graphic and the
% caption, and lengthening the abstract did exactly that. The minipage makes
% image and caption one unbreakable unit; the width is then set so the unit
% still clears the bottom of page 1.
\vspace{2pt}
\noindent\begin{minipage}{\textwidth}
  \centering
  \includegraphics[width=0.84\textwidth]{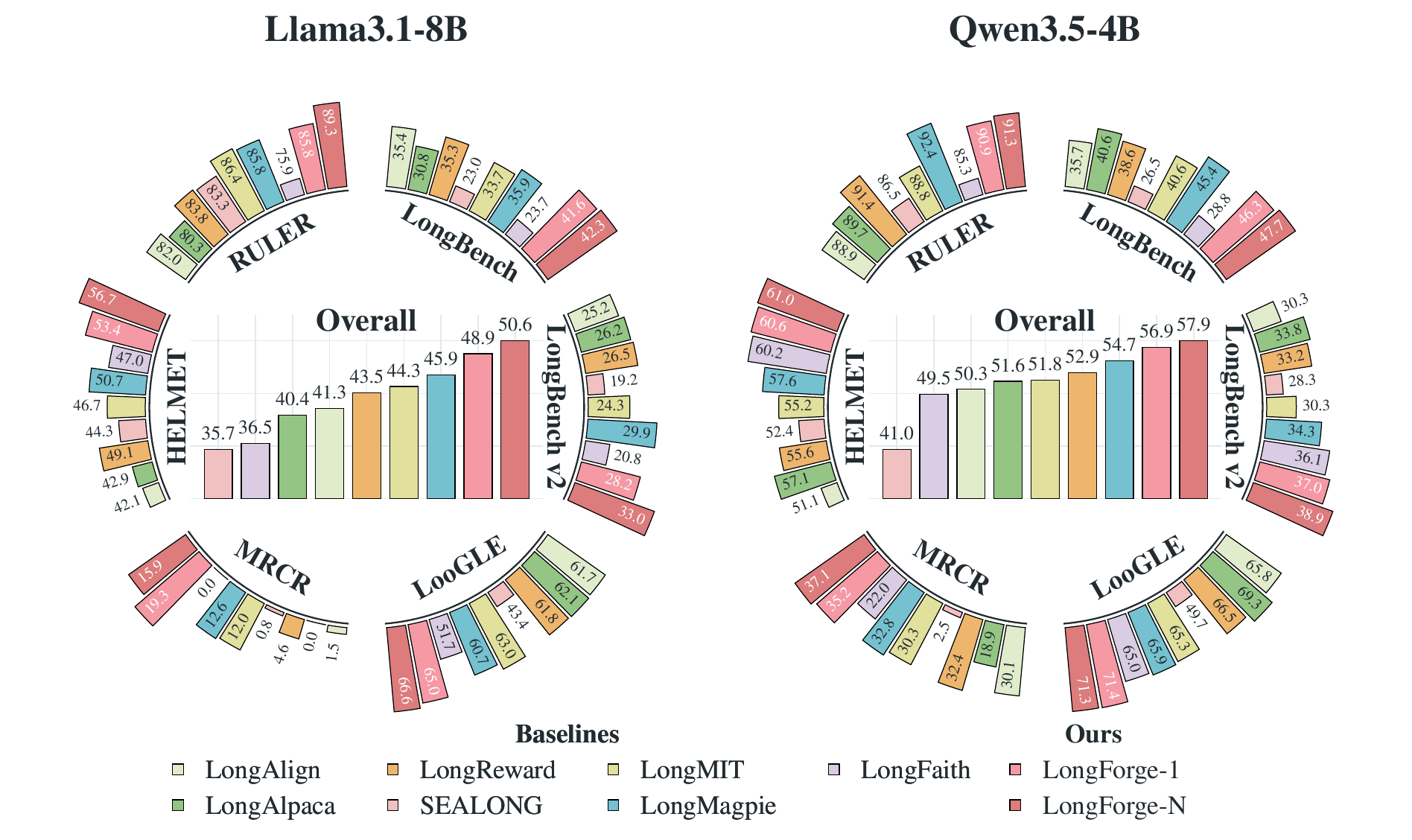}
  \captionof{figure}{Six-benchmark results on the two backbones. Each sector is a benchmark, each bar a training dataset, and the center the six-benchmark average. Our two mixtures lead on both backbones; the numbers are in Table~\ref{tab:main}.}
  \label{fig:main}
\end{minipage}

\section{Introduction}
\label{sec:intro}

\begin{figure}[t]
  \centering
  \includegraphics[width=\textwidth]{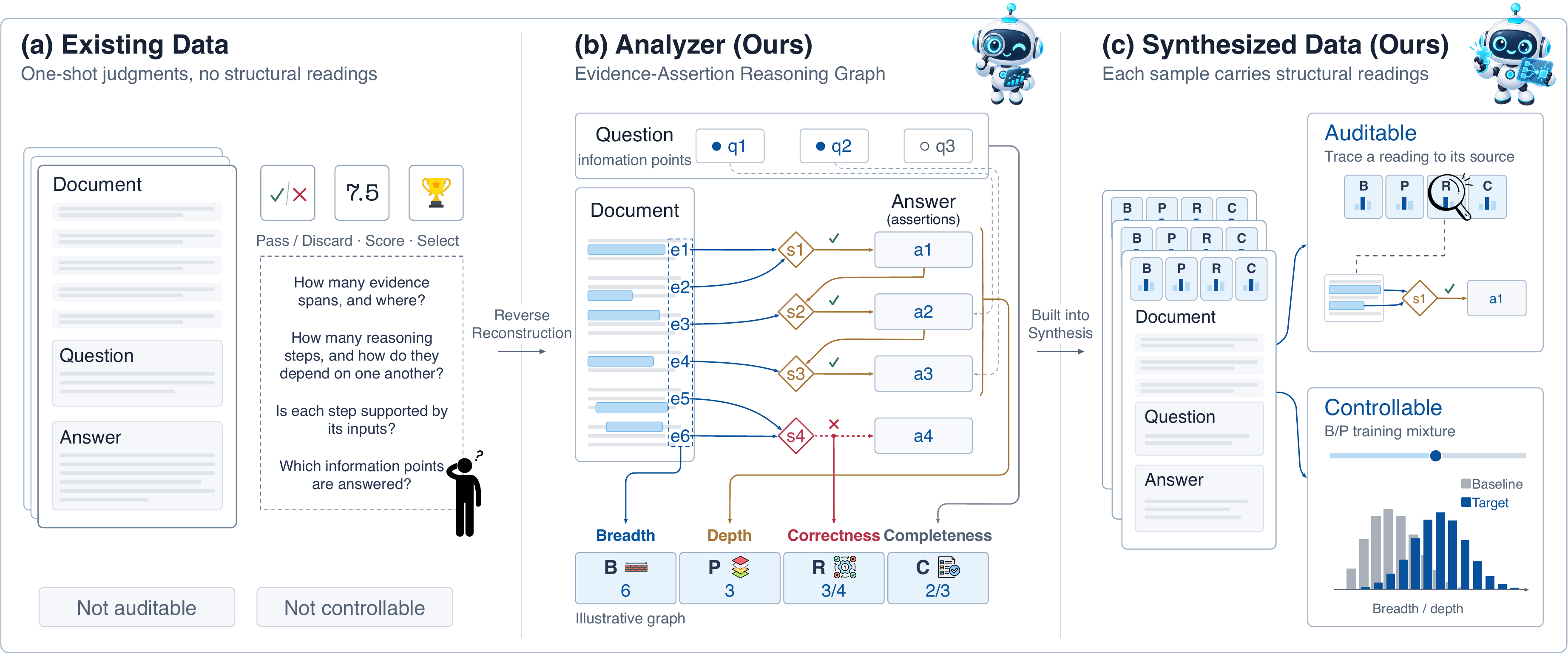}
  \caption{Overview. (a) Existing data keeps only one-shot judgments. (b) The analyzer reconstructs an evidence--assertion reasoning graph and computes four dimensions on it. (c) Each synthesized sample carries these readings and is auditable and controllable.}
  \label{fig:intro}
\end{figure}

The effectiveness of long-context instruction tuning~\citep{chen2024longlora,bai2024longalign} depends heavily on the training data~\citep{chen2026longbenchpro}, yet how to define and measure the quality of long-context instruction-tuning data (hereafter, long-context data) remains open. Existing quality control, whether binary filtering after multi-agent verification~\citep{chen2025longmit}, consistency-based selection~\citep{li2024sealong}, 0--10 scoring~\citep{zhang2025longreward}, or ground-truth citations~\citep{yang2025longfaith}, is a one-shot judgment inside generation that leaves at most a scalar (Figure~\ref{fig:intro}a); selecting samples by their long-range dependencies~\citep{si2025gateau} likewise reduces a sample to one number. A scalar cannot tell which parts of the document an answer uses or how many reasoning steps it writes down, so the data can be neither audited (tracing a reading to its source text and reasoning step) nor controlled (setting the training distribution over these structures).

We define quality on the alignment among the document, the question, and the answer, and build \textbf{LongForge}, which pairs an \emph{analyzer} that measures long-context data with a \emph{synthesizer} that generates it. The analyzer reverse-reconstructs a sample into an \emph{evidence--assertion reasoning graph} (Figure~\ref{fig:intro}b), whose evidence nodes are verbatim document spans, assertion nodes are the answer's own sentences, and information points are the items the question asks for. Programs compute four dimensions on the graph, with the model making only local judgments: \textbf{breadth}, the number of distinct evidence spans used; \textbf{depth}, the longest path from evidence to a terminal assertion after paraphrase contraction; \textbf{correctness}, the share of reasoning steps supported by their inputs; and \textbf{completeness}, the share of information points answered by a non-isolated assertion.

Measuring seven mainstream datasets~\citep{chen2024longlora,bai2024longalign,zhang2025longreward,chen2025longmit,gao2025longmagpie,li2024sealong,yang2025longfaith} reveals two previously unquantified defects. First, breadth barely grows with document length: the median breadth is 5, yet answers cover only 12.1\% of the document on average, and a tenfold longer document raises evidence usage by only about 40\%. Second, depth is low: 43.4\% of the samples write down no explicit intermediate reasoning, and its Pearson correlation with breadth is only 0.070. Existing methods set conditions such as task types or hop counts before generation~\citep{chen2025longmit,sun2026cgmis,yuan2026longcrafter}, yet hops planned on a document graph do not guarantee that the answer writes down the corresponding reasoning, and no method measures these quantities afterwards, so breadth and depth have never been optimization targets. In addition, 23.3\% of the reasoning steps are unsupported and 10.2\% of the samples are incomplete.

The \emph{synthesizer} embeds the analyzer into generation. Correctness and completeness decide whether a sample is usable, so they gate generation, and the failure reasons pinpointed by the analyzer are fed back for re-answering. Breadth and depth measure degree, so they are measured rather than targeted during generation, since targeting them would induce answers that are multi-step only in form, and are mixed when the training set is built. Every sample thus carries its reasoning graph and readings (Figure~\ref{fig:intro}c), making the training distribution directly settable without regeneration.

From one synthesis pool we build two mixtures: \lfone{}, with breadth and depth moderately above the baseline level, and \lfn{}, which organizes the QA pairs of each document into a multi-turn dialogue. On two backbones, they exceed the strongest baseline by 2.2--3.0 and 3.1--4.7 points on average over six long-context benchmarks, whereas raising breadth and depth on the same QA pairs is worth 1.16 points; the gains are robust to the document source and synthesis model, and short-context ability stays on par with the baselines. Our contributions are:
\begin{enumerate}
  \setlength{\itemsep}{0pt}
  \item \textbf{Auditable quality measurement}: an evidence--assertion reasoning graph and an analyzer that compute four traceable quality dimensions on any long-context data.
  \item \textbf{Structural diagnosis of existing data}: across seven mainstream datasets, breadth barely grows with document length and depth is low, the two being nearly orthogonal.
  \item \textbf{Controllable data synthesis}: a synthesizer whose readings make the training distribution settable; its mixtures consistently outperform the strongest baseline.
\end{enumerate}

\section{Related Work}
\label{sec:related}

\paragraph{Long-context data synthesis.}
Most long-context instruction data is synthesized by language models. LongAlpaca~\citep{chen2024longlora} and LongAlign~\citep{bai2024longalign} generate QA pairs over long texts such as papers and books, and LongReward~\citep{zhang2025longreward} follows the LongAlign pipeline; LongMIT~\citep{chen2025longmit} produces multi-hop QA pairs with collaborating agents; LongMagpie~\citep{gao2025longmagpie} lets an aligned model write questions by continuing from a document; SEALONG~\citep{li2024sealong} selects answers by consistency across samples; and LongFaith~\citep{yang2025longfaith} synthesizes faithful reasoning chains from ground-truth answers and citations. Closest to our representation are graph-based methods: CGMIS~\citep{sun2026cgmis} samples multi-hop paths on a concept graph, LongCrafter~\citep{yuan2026longcrafter} builds an evidence graph whose edges are temporal, causal, or coreference links between fragments, and LongMIT assembles multi-document contexts with a coherence graph. These graphs serve synthesis: built before the QA pairs exist, they describe only the document and agree with the generated data by construction. Our graph serves analysis: reconstructed after a sample is given, it adds an assertion layer from the answer and information points from the question, so it can measure how many reasoning steps an answer writes down and whether the answer is complete, and it applies to any existing data.

\paragraph{Data quality assessment.}
For general instruction tuning, quality is assessed on the instruction and the response alone, through curation~\citep{zhou2023lima}, model scoring~\citep{chen2024alpagasus,liu2024deita}, fine-grained tags~\citep{lu2024instag}, instruction-following difficulty~\citep{li2024ifd,li2024superfiltering}, or gradient-based influence~\citep{xia2024less}. Long-context quality further depends on how the answer uses the document. GATEAU~\citep{si2025gateau} selects samples by their long-range dependencies, while LongMIT, SEALONG, LongReward, and LongFaith judge quality inside their synthesis pipelines through binary filtering, consistency, 0--10 scores, or ground-truth citations. All of these yield a scalar or a pass/fail decision that cannot be traced back or used for mixing. Our four dimensions are anchored on the relation between the answer and the document, so each reading can be checked item by item and used directly for mixing.

\section{Analyzer}
\label{sec:analyzer}

Given a long-context QA sample, the analyzer reverse-reconstructs it into an evidence--assertion reasoning graph and computes four dimensions on it (Sections~\ref{sec:graph-def} and~\ref{sec:analyzer-impl}); we then use it to measure seven mainstream datasets (Section~\ref{sec:measurement}).

\subsection{Reasoning Graph and Four Dimensions}
\label{sec:graph-def}

\paragraph{Reasoning graph.}
A sample is a triple $x = (D, Q, A)$, where the document consists of chunks $D = (c_1, \dots, c_M)$, the answer consists of sentences $A = (a_1, \dots, a_K)$, and the question $Q$ literally asks for information points $q_1, \dots, q_L$. The reconstruction is a directed hypergraph
\begin{equation}
  G(x) = (V_E \cup V_A,\; S), \quad V_E \subseteq \mathcal{E}(D), \quad V_A = \{a_1, \dots, a_K\},
  \label{eq:graph}
\end{equation}
where $\mathcal{E}(D) = \{(m, i, j) : 1 \le m \le M,\, 1 \le i \le j \le |c_m|\}$ is the set of spans ``sentences $i$ to $j$ of chunk $c_m$,'' and each hyperedge $s = (\mathrm{In}(s) \to \mathrm{out}(s)) \in S$, with $\mathrm{In}(s) \subseteq V_E \cup V_A$ and $\mathrm{out}(s) \in V_A$, is a reasoning step in which evidence spans and/or preceding assertions jointly support one assertion. Evidence nodes are thus verbatim source text, each used by at least one step, and the assertions are exactly the answer sentences. These two constraints keep the graph faithful to the sample: the reconstruction can neither introduce evidence absent from the document nor build a chain deeper or more complete than the answer, so the graph reflects the sample rather than the ability of the analyzer's model. Assertions output by no step are \emph{isolated}, $V_A^{\mathrm{iso}} = V_A \setminus \{\mathrm{out}(s) : s \in S\}$, and correspond to answer content that cannot be derived from the document.

\paragraph{Four dimensions.}
Let $\sigma(s) \in \{0, 1\}$ indicate whether the inputs of step $s$ support its output, and $\mathrm{cov}(q_\ell) \in \{0,1\}$ whether $q_\ell$ is answered by a non-isolated assertion. The four dimensions, breadth $B$, depth $P$, correctness $R$, and completeness $C$, are
\begin{equation}
  B(x) = |V_E|, \quad P(x) = \max_{v \in F} \mathrm{len}_{\tilde{G}}(v), \quad R(x) = \frac{1}{|S|} \sum_{s \in S} \sigma(s), \quad C(x) = \frac{1}{L} \sum_{\ell=1}^{L} \mathrm{cov}(q_\ell),
  \label{eq:dims}
\end{equation}
where $F \subseteq V_A \setminus V_A^{\mathrm{iso}}$ is the set of terminal assertions that directly answer the question, and $\mathrm{len}_{\tilde{G}}(v)$ is the length, in steps, of the longest path from evidence to $v$. Depth is computed on the quotient graph $\tilde{G}$ obtained by paraphrase contraction: a step whose output merely rewords its inputs derives nothing new, so an assertion paraphrasing evidence is treated as that evidence and one paraphrasing a preceding assertion is merged with it. Without this contraction, depth would merely count answer sentences. The other three dimensions are computed on $G(x)$. Since the four dimensions read distinct structural positions (node counts, path lengths, edge judgments, and question-side coverage), we report them separately rather than combining them.

Two properties follow. Because assertions come only from answer sentences, depth measures how many steps the answer writes down explicitly, not how hard the question is, and $P(x) \le |V_A \setminus V_A^{\mathrm{iso}}| \le K$; unsupported steps still count toward depth, their validity being measured by correctness. Because the graph contains only what the answer uses and writes, omissions are invisible to breadth, depth, and correctness; completeness detects them by taking the question's information points as the denominator.

\subsection{Analyzer Implementation}
\label{sec:analyzer-impl}

\begin{figure}[t]
  \centering
  \includegraphics[width=\textwidth]{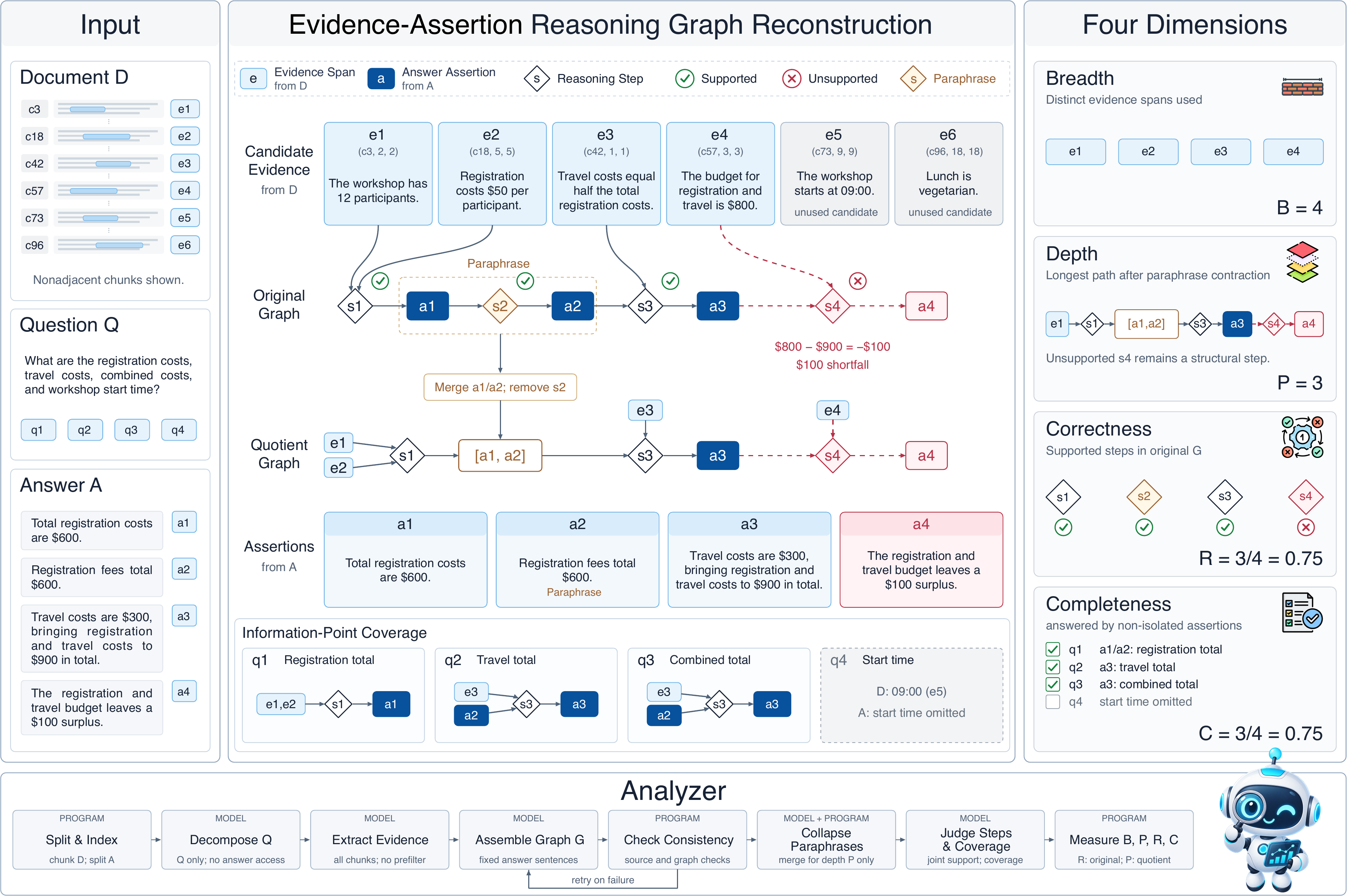}
  \caption{The analyzer pipeline (bottom) and an example (top) with $B = 4$, $P = 3$, $R = 3/4$, and $C = 3/4$.}
  \label{fig:analyzer}
\end{figure}

The pipeline (Figure~\ref{fig:analyzer}) alternates program steps and calls to a single model, DeepSeek-V4.1-Flash~\citep{deepseekv41flashreport}. Question decomposition, graph assembly, paraphrase judgment, and information-point coverage require integrating more context and run in thinking mode; evidence extraction and step judgment, the most frequent calls, run without thinking and with greedy decoding for reproducibility. We detail each stage below.

\paragraph{Split \& Index.}
The analyzer first uses a program to split the document into chunks at paragraph and sentence boundaries (at most 1{,}200 characters per chunk) and numbers every sentence, so that evidence can be located precisely by sentence indices; it then splits the answer into sentences as assertions $a_1, \dots, a_K$.

\paragraph{Decompose $Q$.}
The analyzer asks the model to decompose the question into information points $q_1, \dots, q_L$. The model sees only the question, not the answer, so that the information points are not influenced by the answer.

\paragraph{Extract Evidence.}
The analyzer calls the model on each chunk in parallel, asking it to find sentence ranges in the chunk that support an assertion or respond to an information point as candidate evidence and to output only sentence indices; a program then retrieves the source text by index, ensuring that evidence is verbatim. The analyzer traverses all chunks instead of first narrowing the scope by retrieval, so that evidence used by the answer is not undercounted because of retrieval misses; the cost is about 100 model calls per baseline sample on average. The analyzer does not pass evidence that only responds to an information point without supporting any assertion to the assembly stage, to avoid inflating breadth.

\paragraph{Assemble Graph $G$.}
The analyzer gives all assertions and candidate evidence to the model, asks it to connect the assertions into a graph by derivation relations, and to mark isolated assertions, terminal assertions, and redundant evidence to be removed.

\paragraph{Check Consistency.}
The analyzer then checks the graph structure with a program so that all four dimensions are computed on a valid graph: whether all indices are valid, whether each assertion is either output by exactly one step or marked isolated, whether the graph is acyclic, and whether every terminal assertion is reachable from evidence. If the check fails, the analyzer feeds the failure reasons back to the model and asks it to reassemble, at most three times; samples that still fail are marked as degenerate samples.

\paragraph{Collapse Paraphrases.}
The analyzer judges in two tiers whether a step is merely a paraphrase. If more than 85\% of the content words of the output assertion already appear in its input evidence, the analyzer classifies it as paraphrasing evidence by program; paraphrases of preceding assertions often change the wording and cannot be distinguished from genuine reasoning by surface overlap alone, so the analyzer leaves them to the model.

\paragraph{Judge Steps \& Coverage.}
The analyzer asks the model to judge, step by step, whether the inputs derive the output, which gives $\sigma(s)$; unparsable judgments are treated as unsupported to avoid inflating correctness. For each information point, the analyzer asks the model to identify the assertions that answer it, or to judge that none exists, and then removes isolated assertions among them by program.

\paragraph{Measure $B$, $P$, $R$, $C$.}
Finally, the analyzer computes the four dimensions by program according to the definitions in Section~\ref{sec:graph-def}; the model provides only the local judgments above and never outputs the value of any dimension directly. The analyzer also stores the reasoning graph, together with the judgment of every step, alongside the sample, so that every reading can be traced back to specific evidence spans and reasoning steps.

\subsection{Measuring Existing Baselines}
\label{sec:measurement}

To characterize the structural form of existing long-context data, we measure the seven datasets of Section~\ref{sec:related} with the analyzer, drawing 1{,}000 random samples from each. All self-checks of the reconstruction pass: the depth of every sample does not exceed its number of non-isolated assertions, all chunks are traversed, and degenerate samples account for 2.4\%; after excluding them, 6{,}834 samples enter the statistics. We report the four dimensions in turn, then decompose the failure modes, and finally report a human audit.

\begin{figure}[t]
  \centering
  \includegraphics[width=\textwidth]{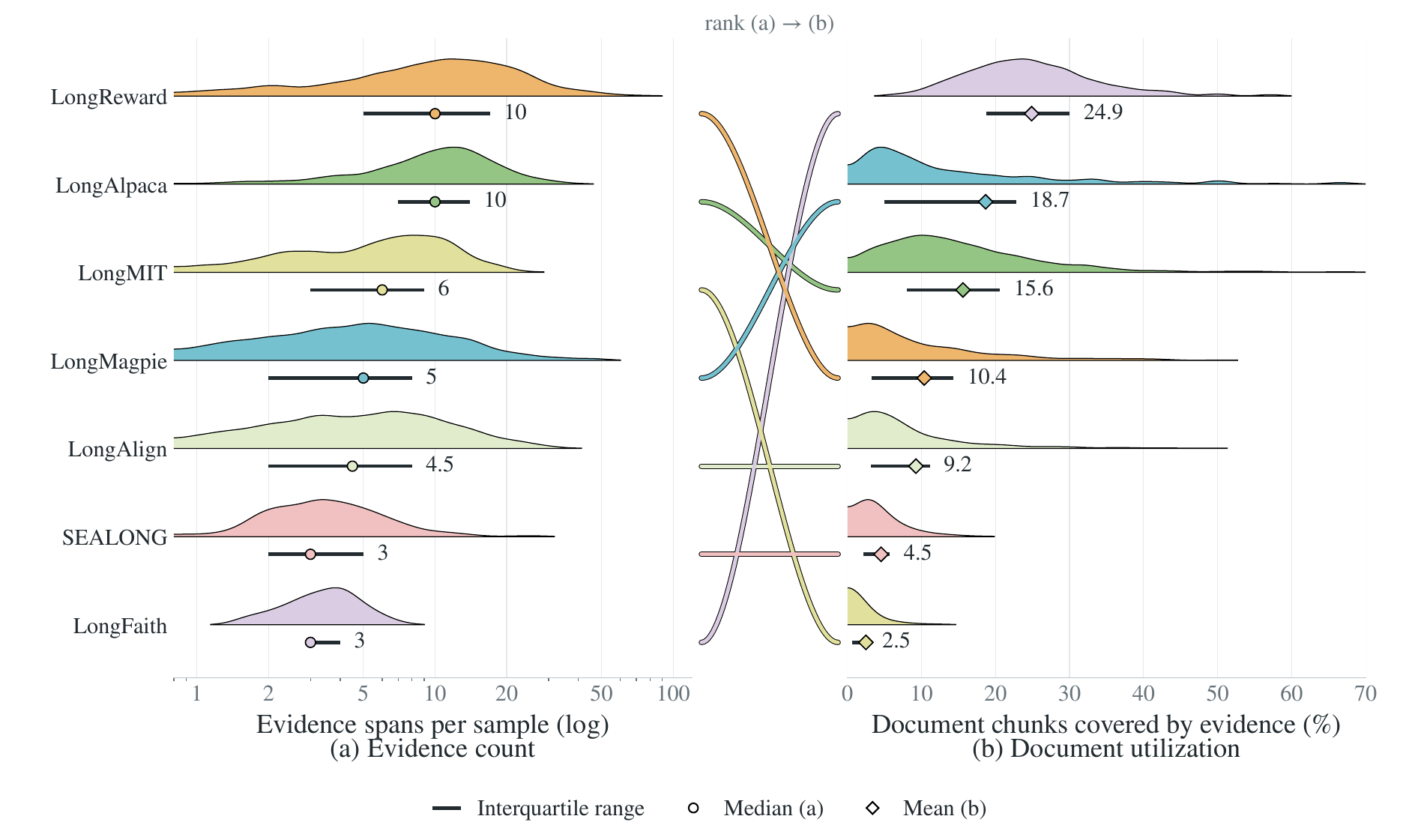}
  \caption{Per-sample evidence count (a, log scale) and document utilization (b). Each row is a baseline; the lines in the middle connect the positions of the same baseline in the two panels.}
  \label{fig:breadth}
\end{figure}

\begin{figure}[t]
  \centering
  \includegraphics[width=\textwidth]{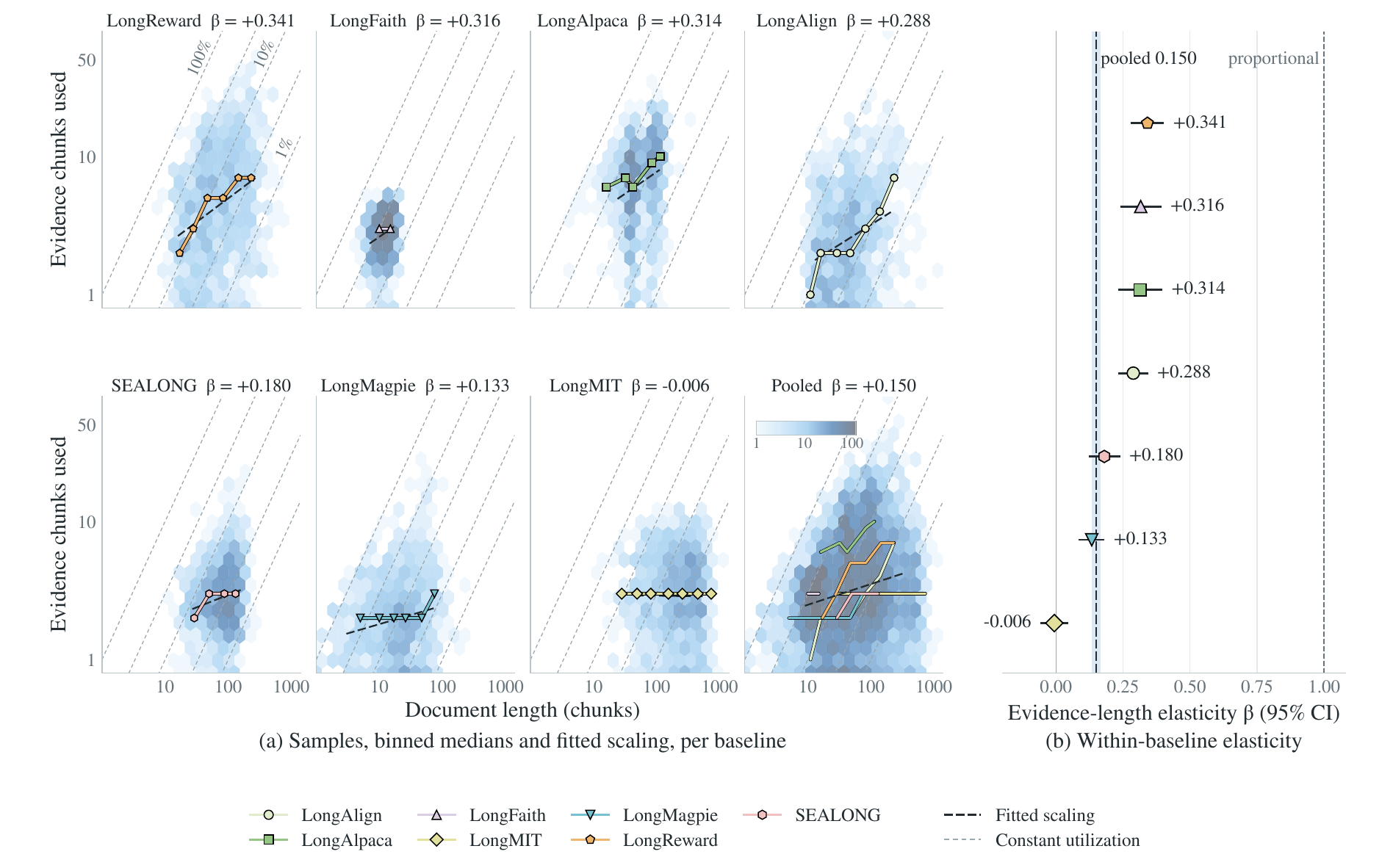}
  \caption{Evidence-length elasticity. (a) Per-baseline scaling; the gray dashed lines are iso-utilization lines, the colored line is the median by document-length bin, and the black dashed line is the log--log fit, whose slope is the elasticity $\beta$. (b) $\beta$ with 95\% confidence intervals.}
  \label{fig:elasticity}
\end{figure}

\paragraph{Breadth: the number of evidence spans is not small, yet it barely grows with document length.}
The median breadth is 5, so the number of evidence spans itself is not small, but it covers only a small part of the document: \emph{document utilization}, i.e., the share of chunks that yield evidence among all chunks, is only 12.1\% overall. The rankings by evidence count and by utilization barely correspond (Figure~\ref{fig:breadth}): LongFaith, with the fewest evidence spans, has the highest utilization because its documents are the shortest; LongMIT, with the longest documents, has the lowest utilization, only 0.025. The problem is thus not that there are too few evidence spans, but that the amount of evidence barely grows as documents get longer. To quantify this, we estimate the \emph{evidence-length elasticity} $\beta$, the log--log regression slope of the number of chunks yielding evidence on the number of chunks in the document. It ranges from $-0.006$ (LongMIT) to 0.341 (LongReward), with a pooled value of $0.150 \pm 0.008$, far below the value of 1 required for proportional growth (Figure~\ref{fig:elasticity}): a tenfold longer document raises evidence usage by only about 40\% and cuts utilization to about one-seventh. Existing data thus demonstrates the use of only a few fragments of a document regardless of its length, and the longer the document, the wider this gap.

\begin{figure}[t]
  \begin{minipage}[t]{0.48\textwidth}
    \centering
    \includegraphics[width=\linewidth]{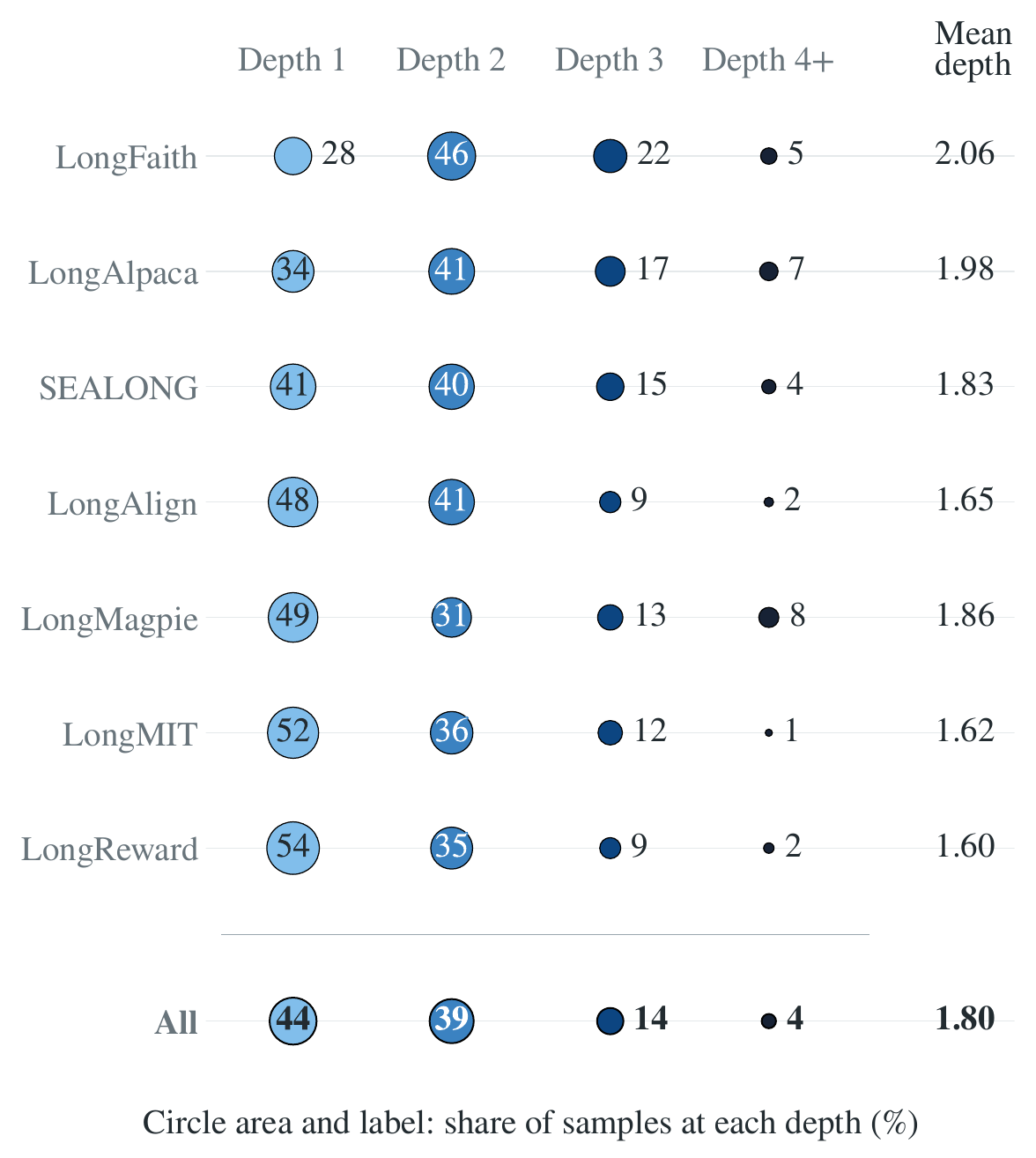}
    \caption{Depth distribution per dataset; the last row pools all samples. The area and value of each circle give the share of samples at that depth ($P = 1$ means no explicit intermediate reasoning), and the right column gives the mean depth.}
    \label{fig:depth}
  \end{minipage}\hfill
  \begin{minipage}[t]{0.48\textwidth}
    \centering
    \includegraphics[width=\linewidth]{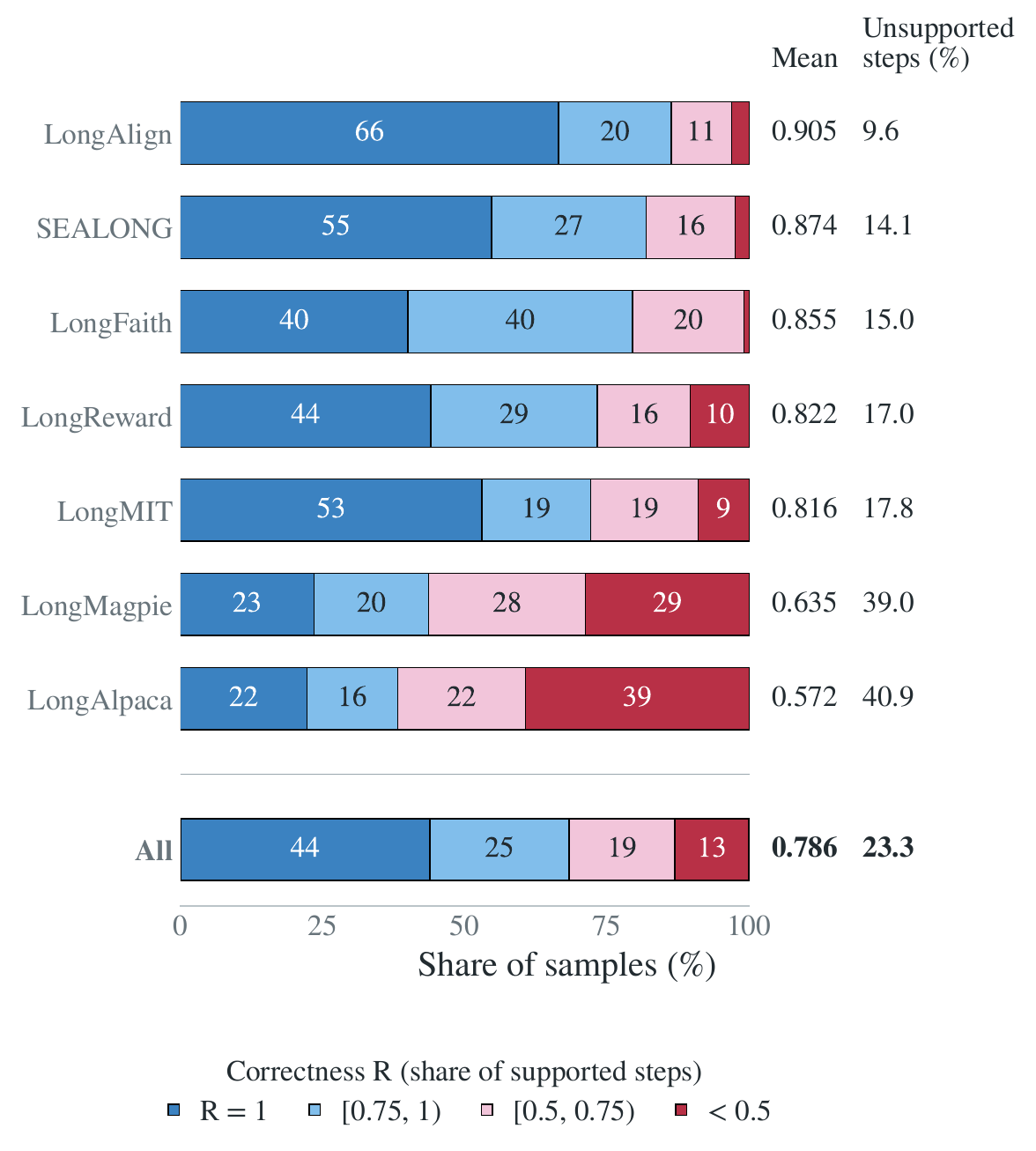}
    \caption{Correctness composition per dataset; the last row pools all samples. Each bar is split by $R$ into four segments ($R = 1$, $[0.75, 1)$, $[0.5, 0.75)$, $R < 0.5$), and the right columns give the mean and the share of unsupported steps.}
    \label{fig:correctness}
  \end{minipage}
\end{figure}

\paragraph{Depth: four in ten samples write down no explicit intermediate reasoning.}
43.4\% of the samples have depth 1, i.e., no explicit intermediate reasoning, 82.2\% have at most two steps, and the mean depth is 1.798 (Figure~\ref{fig:depth}). Paraphrase contraction is indeed necessary: 486 samples have depth at least 3 before contraction and 1 after it, their nominal multi-step form being a reworded repetition of their inputs. The Pearson correlation between depth and breadth is only 0.070 at the sample level, and 0.025 to 0.258 within individual datasets (Appendix~\ref{app:joint}), so the two are axes that do not determine each other: raising evidence usage does not raise depth. LongFaith, which has the highest depth (72.5\% multi-step), is the only one of the seven that keeps complete reasoning chains during construction, confirming that the dimension measures how many steps the answer writes down. High depth does not mean higher reliability, however: the rank correlation between depth and correctness is $-0.253$ (all pairwise correlations are in Appendix~\ref{app:correlation}), i.e., the longer the written reasoning, the more likely one step does not hold; weighting the four dimensions into a single score would erase this relation, which is why we report them separately.

\paragraph{Correctness: nearly a quarter of the reasoning steps do not follow from their inputs.}
The overall mean correctness is 0.786, but only 43.8\% of the samples have every step supported (Figure~\ref{fig:correctness}); counted by steps, 23.3\% of the reasoning steps are unsupported. The dataset means range from 0.905 (LongAlign) to 0.572 (LongAlpaca), and the differences lie mainly in the severity of failure: among the five datasets with higher means, samples with $R < 0.5$ make up at most about one-tenth, and failures are mostly individual unsupported steps; whereas in LongMagpie and LongAlpaca, 29\% and 39\% of the samples have more than half of their reasoning steps unsupported. A considerable part of the reasoning in existing data thus demonstrates unfounded inference, and in most datasets the failures fall on individual steps in the chain and can only be found by step-by-step checking.

\begin{figure}[t]
  \centering
  \includegraphics[width=\textwidth]{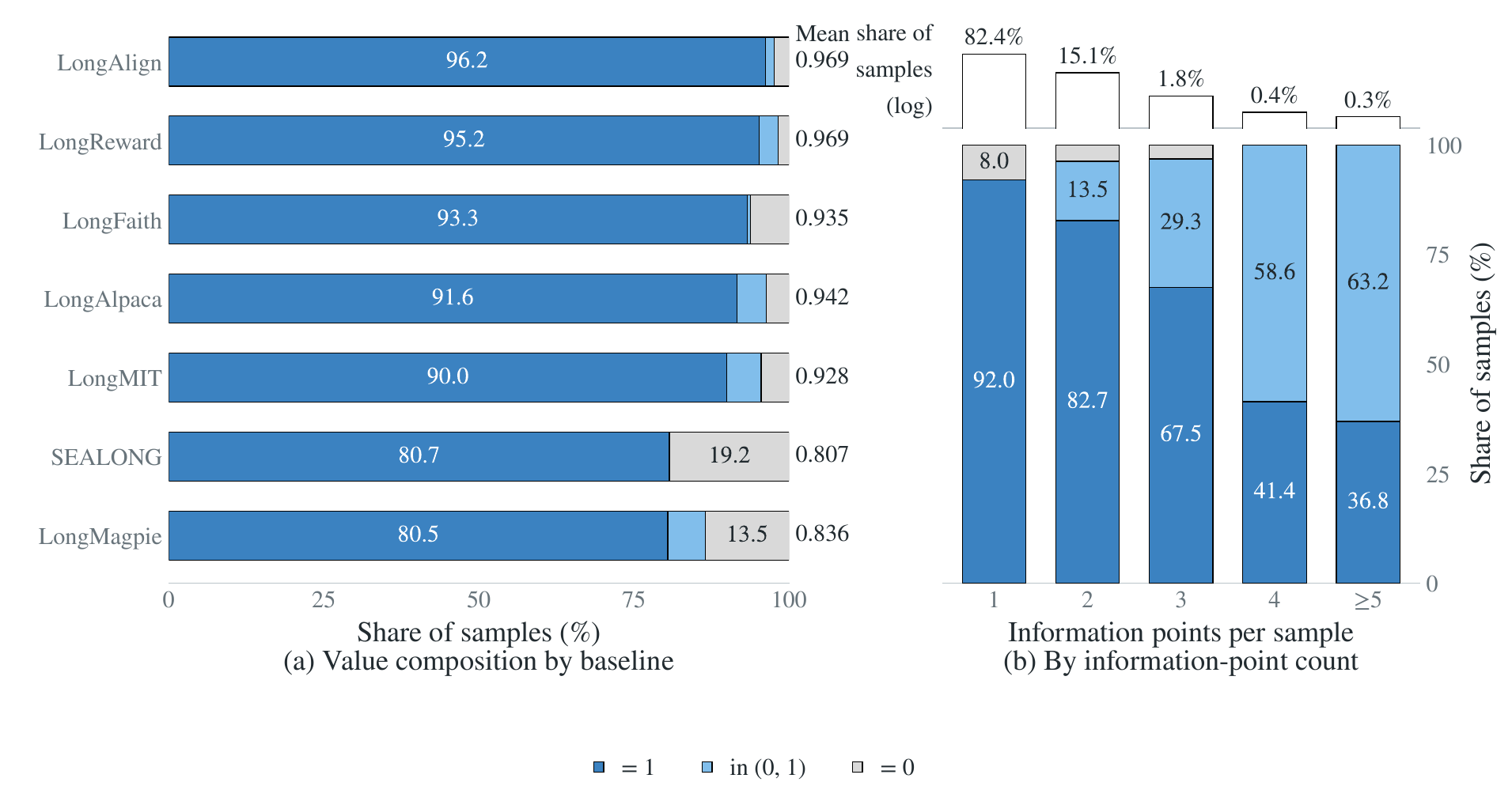}
  \caption{Composition of completeness values. (a) Share of samples with completeness $=1$, in $(0, 1)$, and $=0$ in each dataset; the right column gives the mean. (b) Value composition stratified by the number of information points in the question; the top strip gives the share of samples in each stratum (log scale).}
  \label{fig:completeness}
\end{figure}

\paragraph{Completeness: most questions ask for a single information point, so the values sit at the two ends.}
The overall mean completeness is 0.913, but the values fall almost entirely at the two ends: 89.8\% of the samples take 1 and 7.2\% take 0, with only 3.0\% in the open interval (Figure~\ref{fig:completeness}a). The differences among datasets lie mainly in the share of zeros: SEALONG and LongMagpie, with the lowest means, have 19.2\% and 13.5\% of their samples answering none of what the question asks for. This shape stems from the questions themselves: 82.4\% of the samples have only one information point, and with a denominator of 1, completeness can only be 0 or 1. The more information points, the larger the share of values in the open interval, rising from 13.5\% with two points to over half with four or more (Figure~\ref{fig:completeness}b). Most questions in existing long-context data ask for one thing, so ``whether the answer is complete'' degenerates into a binary judgment, and high completeness does not mean that answers are comprehensive.

\begin{figure}[t]
  \centering
  \includegraphics[width=0.6\textwidth]{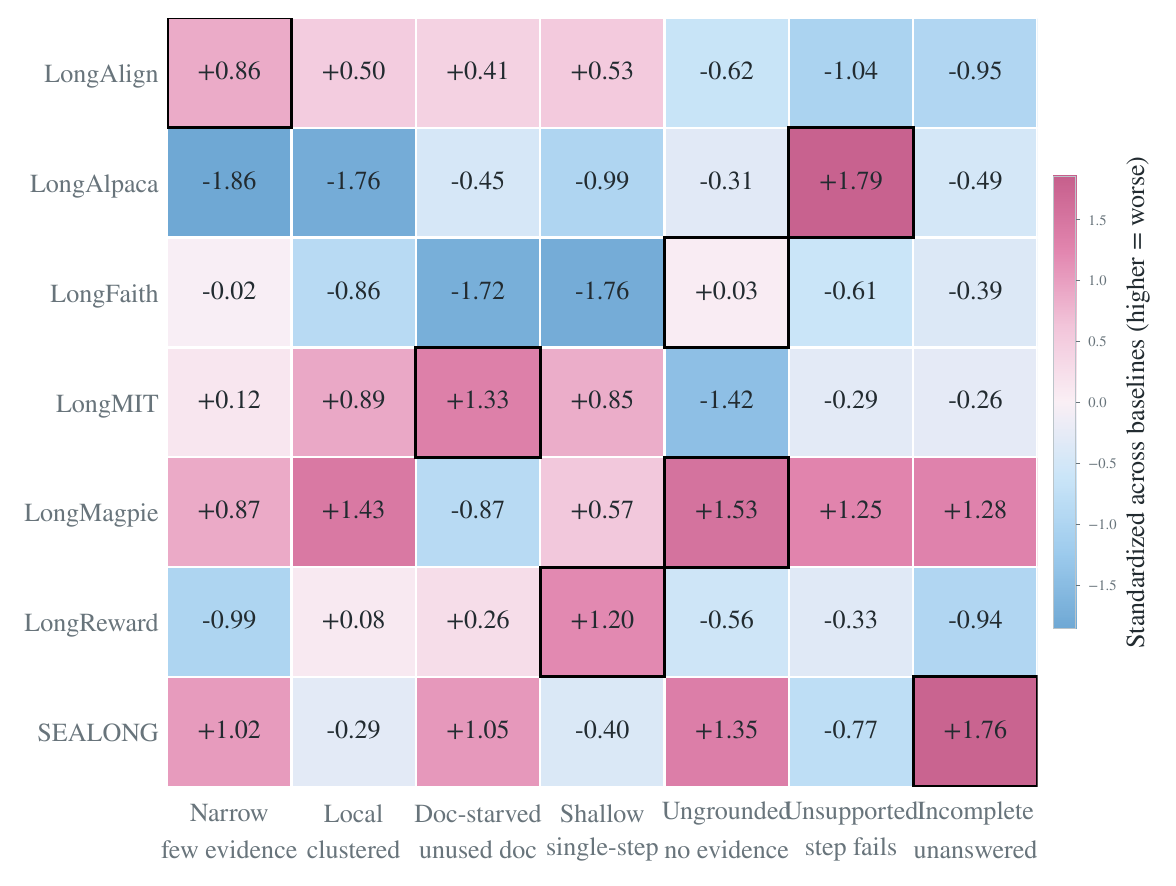}
  \caption{Failure-mode matrix. Each row is a dataset and each column a failure axis; each cell is the value standardized across baselines within the axis (higher = worse), and the black box marks the dataset's dominant defect.}
  \label{fig:failure}
\end{figure}

\paragraph{Failure modes: the dominant defect differs across the seven datasets.}
To examine whether the datasets fail in the same way, we decompose failures into seven axes, namely Narrow, Local, Doc-starved, Shallow, Ungrounded, Unsupported, and Incomplete, and standardize each axis across datasets; the axis definitions are given in Appendix~\ref{app:failure}. The dominant defects of the seven datasets are spread over six different axes and span all four dimensions (Figure~\ref{fig:failure}); conversely, their best-performing axes are spread over five axes, and no dataset performs poorly on all axes. The analyzer can thus distinguish the specific weaknesses of different datasets.

\paragraph{Human audit: the readings can be checked item by item and agree closely with human judgment.}
Auditability is the property that distinguishes our measurement from earlier quality assessment, and one human audit tests both this property and the reliability of the readings. Two auditors experienced in long-context annotation independently check the four dimensions on 346 samples stratified by dataset, evidence by evidence and step by step, rating each dimension as consistent, minor deviation (a small discrepancy that does not cross a qualitative boundary), or major deviation; we report the stricter auditor below, with details in Appendix~\ref{app:audit}.

\begin{figure}[!ht]
  \centering
  \includegraphics[width=\textwidth]{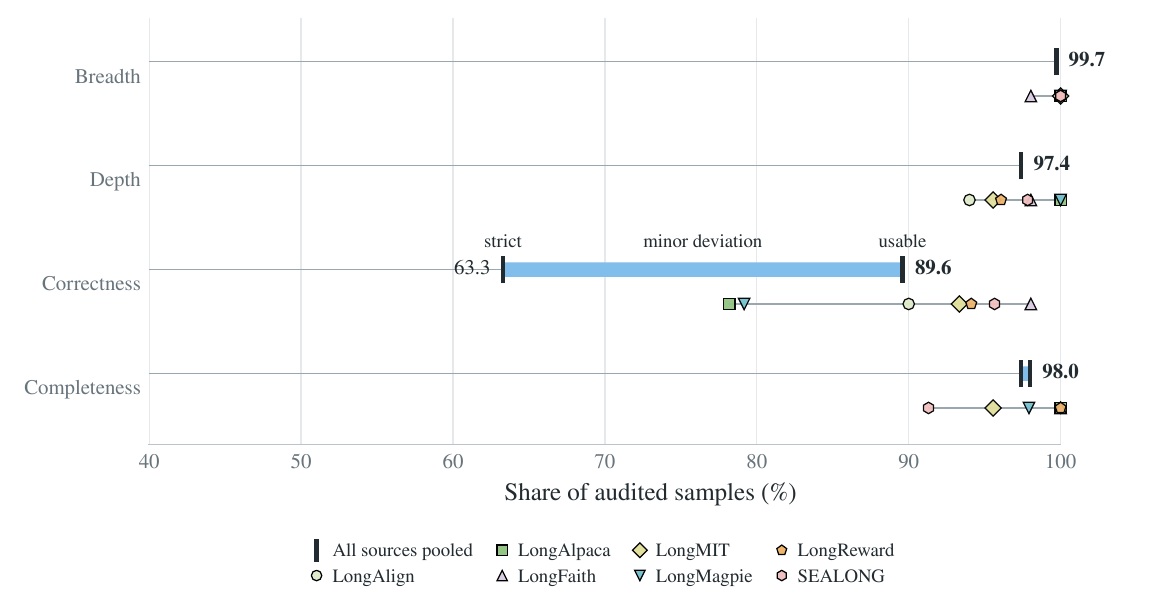}
  \caption{Human-audit agreement: strict agreement and usable rates, pooled (top) and per source (bottom).}
  \label{fig:audit}
\end{figure}

Such item-by-item checking is possible only because every evidence span, assertion, and step judgment is stored in the reasoning graph, so a disagreement can be located at a specific evidence span or reasoning step; faced with a scalar, an auditor could only agree or disagree with the sample as a whole. Measured by the usable rate (the share rated consistent or minor deviation), the four dimensions reach 99.71\% for breadth, 97.40\% for depth, 89.60\% for correctness, and 97.98\% for completeness (Figure~\ref{fig:audit}). Correctness is lower because it aggregates step judgments: per-step agreement is 0.928, on the same order as the other three dimensions, but with a median of 6 steps per sample, small disagreements accumulate along the chain; even so, only 10.4\% are major deviations.

\section{Synthesizer}
\label{sec:synthesizer}

The measurements of Section~\ref{sec:analyzer} show that existing long-context data has structural defects on all four dimensions. The question that follows is whether breadth and depth can be raised while correctness and completeness are guaranteed, and what such data brings. This section designs a synthesizer that constructs such data with the same document sources as the baselines, so that the question can be answered in a controlled manner: Section~\ref{sec:synth-impl} describes the implementation and Section~\ref{sec:synth-dist} analyzes the data distributions of the two mixtures.

\subsection{Synthesizer Implementation}
\label{sec:synth-impl}

\begin{figure}[t]
  \centering
  \includegraphics[width=\textwidth]{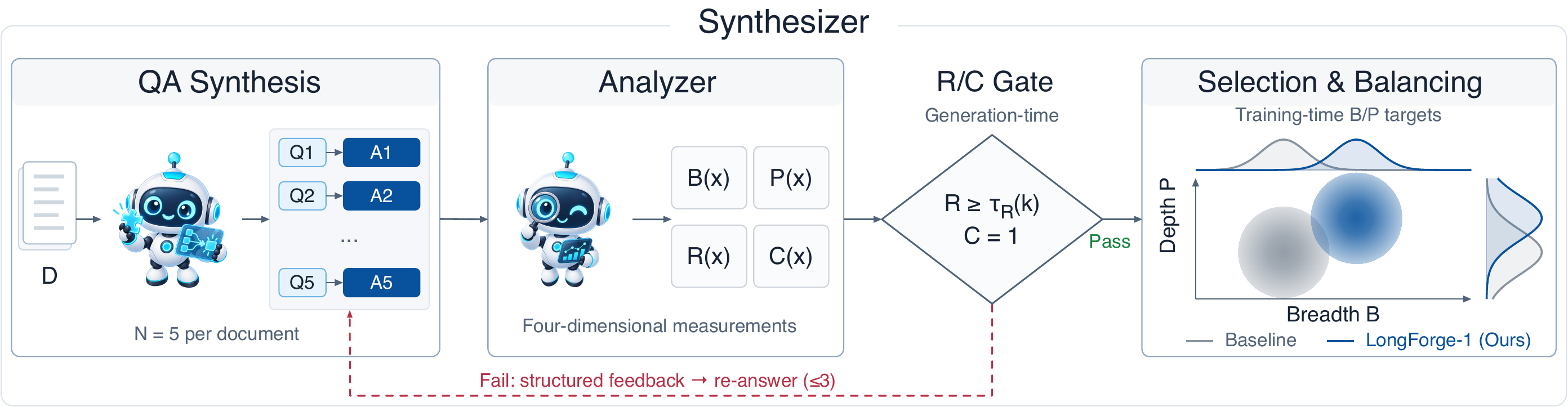}
  \caption{The synthesizer pipeline. Left: question generation and answering. Middle: measurement and gating by the analyzer, with failing QA pairs re-answered from their failure reasons. Right: the distribution before and after mixing by breadth and depth when the training set is built.}
  \label{fig:synthesizer}
\end{figure}

The design is shown in Figure~\ref{fig:synthesizer}. Its core is to apply the four dimensions at two different places: correctness and completeness, which decide whether a sample is usable, act as a gate at generation time, whereas breadth and depth, which measure degree, are deferred to the mixture built at training-set construction time.

\paragraph{QA synthesis.}
QA generation is deliberately minimal, so that differences from the baselines come mainly from the analyzer-driven gating and mixing: documents are used as is, question generation and answering each use a single simple prompt, and neither task types nor reasoning paths are specified in advance. For each document, the synthesizer gives the model the full source text and generates five questions at once. The prompt states only three requirements: answering must combine evidence from several places in the document, answering must require multiple reasoning steps, and the questions on one document must rely on evidence as different as possible. The last requirement targets document utilization: the evidence used by a single answer barely grows as the document gets longer, so raising utilization requires several questions falling on different regions. The answering prompt specifies no output format and does not ask the model to ``write out the reasoning process,'' to avoid inducing reworded restatements. The full prompts are in Appendix~\ref{app:prompts}.

\paragraph{Measurement, gating, and re-answering.}
Once a QA pair is generated, the synthesizer passes it to the analyzer to rebuild the reasoning graph and compute the four dimensions; the analyzer is exactly the one used to measure the baselines, so the readings of synthesized and baseline data are directly comparable. If correctness $R$ falls below a threshold $\tau_R(k)$ or completeness $C$ does not reach 1, the synthesizer feeds the failure reasons read from the graph, i.e., which steps are unsupported and which information points are unanswered, back to the model for re-answering, at most three times; pairs that still fail do not enter the training set. What is fed back is not a scalar signal such as ``the answer is wrong,'' but the specific step and information point at fault, a capability that comes precisely from representing quality as structure rather than as a scalar.

The threshold $\tau_R(k)$ is not uniformly set to the full value but tiered by the number of reasoning steps $k = |S|$: 1.00 for $k \le 2$, 0.85 for $3 \le k \le 5$, and 0.75 for $k \ge 6$, i.e., a long chain is allowed one questionable step. This is because correctness aggregates step judgments over the whole reasoning chain, and small disagreements in those judgments accumulate along the chain (Section~\ref{sec:measurement}); if one questionable step vetoed the whole sample, longer chains would be rejected more often by mistake, while long chains are exactly the samples we need.

\paragraph{Two mixtures.}
Every QA pair in the synthesis pool carries readings of the four dimensions, so a training set with any desired distribution can be built directly. We design two mixtures:
\begin{itemize}
  \setlength{\itemsep}{0pt}
  \item \textbf{\lfone{}} raises breadth and depth moderately above the baseline level. Taking one QA pair per document, it samples the breadth tiers $B < 6$, $6 \le B < 12$, $12 \le B < 20$, and $B \ge 20$ with quotas of 20\%, 40\%, 30\%, and 10\%, and requires $P \ge 2$ in the top two tiers. The lower tiers do not constrain depth, so as to keep the ``few evidence spans, single step'' form that dominates existing data; the top two tiers require depth of at least 2, so that samples that spread out many evidence spans but reason only one step are not taken as reasoning. Since breadth and depth are nearly orthogonal, no separate quota is set for depth.
  \item \textbf{\lfn{}} aims at document utilization. A single QA pair can cover only a limited share of a document, and question generation already requires the questions on one document to rely on different evidence; \lfn{} therefore organizes all QA pairs of a document that pass the gate into a multi-turn dialogue, with the document appearing only in the first turn and one question given in each turn thereafter. One training sample thus covers the union of the evidence of all turns, and document utilization rises accordingly.
\end{itemize}

\subsection{Data Distributions of the Two Mixtures}
\label{sec:synth-dist}

The documents actually used for synthesis come from a mixed pool formed by the baseline datasets, so that the document sources match the baselines. We draw 4{,}096 documents at random, generate five questions each, and obtain 20{,}480 QA pairs, 99.21\% of which pass the gate (per-round pass rates, failure reasons, and call cost are in Appendix~\ref{app:synth-stats}).

\begin{figure}[t]
  \centering
  \includegraphics[width=\textwidth]{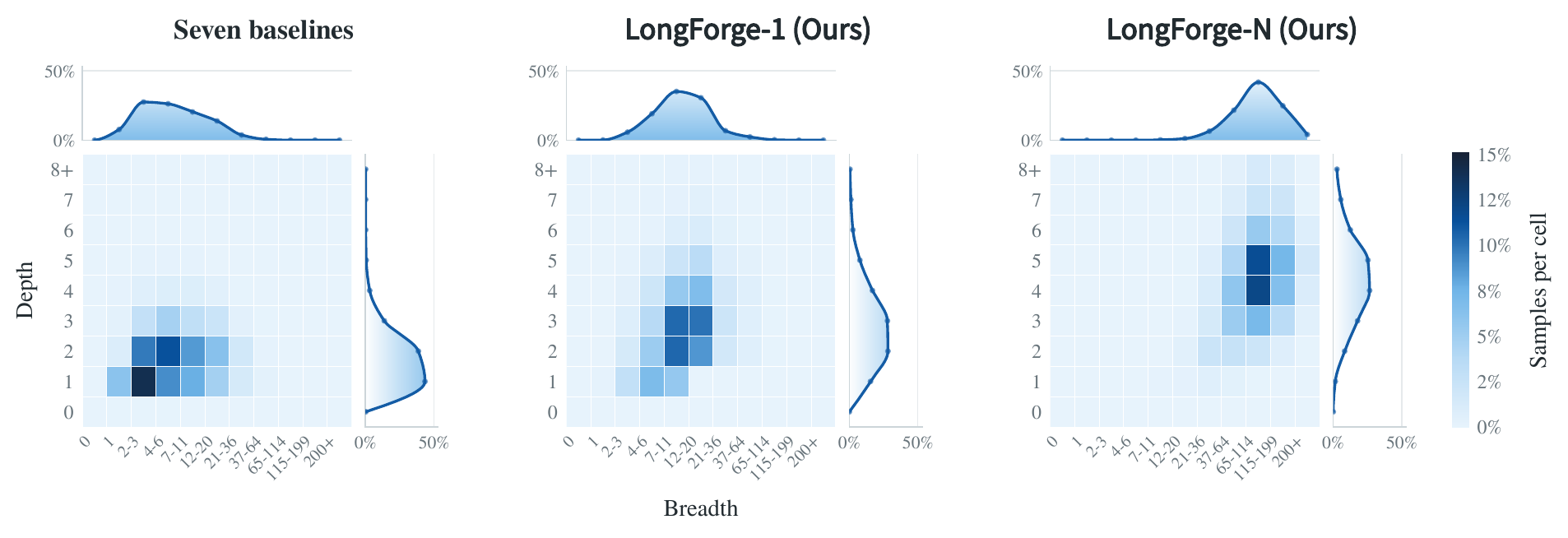}
  \caption{Joint distribution of breadth and depth per training sample. Each cell is the share of samples falling in it, with the marginal distributions on the top and right; the three panels share one set of bins and one probability scale. The version with the seven baselines shown separately is Figure~\ref{fig:by-baseline} in Appendix~\ref{app:joint}.}
  \label{fig:joint}
\end{figure}

This section compares the two mixtures with the seven baselines on the four dimensions; the unit of measurement is one training sample. A baseline or \lfone{} sample contains a single QA pair, whereas an \lfn{} sample consists of several, and each dimension is aggregated across turns according to the position it reads on the reasoning graph: breadth takes the union of the evidence of all turns, depth the maximum over turns, correctness the share of supported steps among all reasoning steps of all turns, and completeness requires all information points of all turns to be answered. Neither breadth nor depth is summed, to avoid double-counting shared evidence or fabricating a path that no answer actually follows; after this aggregation, the three can be compared on the same scale. Figure~\ref{fig:joint} and Table~\ref{tab:dist} give the comparison.

\begin{table}[t]
  \centering
  \small
  \setlength{\tabcolsep}{8pt}
  \begin{tabular}{@{}l!{\color{black!40}\vrule}ccc@{}}
    \toprule
    & Seven baselines & \lfone{} & \lfn{} \\
    \midrule
    Breadth $B$ & 7.18 & 12.2 & 95.5 \\
    Depth $P$ & 1.798 & 2.89 & 4.42 \\
    Single-step rate & 0.434 & 0.154 & 0.015 \\
    Correctness $R$ & 0.786 & 0.928 & 0.918 \\
    Completeness $C$ & 0.913 & 1.000 & 1.000 \\
    Document utilization & 0.121 & 0.102 & 0.519 \\
    \bottomrule
  \end{tabular}
  \caption{Data distributions of the two mixtures alongside the seven baselines. The single-step rate is the share of samples with depth 1; all other rows are sample means.}
  \label{tab:dist}
\end{table}

\lfone{} achieves the intended moderate increase: mean breadth 12.2, 1.7 times the baselines; mean depth 2.89, 1.6 times; and a single-step rate dropping from 0.434 to 0.154. Every reading varies by less than 0.15 across the three samplings, so the distribution obtained by mixing on measured values is stable and reproducible, which is controllability in its most direct form.

The shift of \lfn{} comes from its unit of organization: mean breadth reaches 13.3 times the baselines, mean depth 2.5 times, single-step samples almost vanish (0.015), and document utilization rises to 0.519. This follows from the evidence of different questions barely overlapping: the median ratio of the evidence union to the summed evidence counts over the five questions on a document is 0.965, so the union accumulates almost linearly. By contrast, the document utilization of \lfone{} (0.102) is of the same order as the baselines (0.121), confirming that a single QA pair can cover only a limited share of a document and that raising utilization requires several complementary QA pairs.

Both mixtures are also clearly above the baselines in correctness and completeness, a direct result of the generation-time gate: mean correctness rises from 0.786 to 0.928 and 0.918, and mean completeness from 0.913 to 1.000, with the information points of every sample answered. The synthesized questions also ask for more than the baselines: the mean number of information points is 1.78 and only 57.2\% of the questions have a single information point, against 1.22 and 82.4\% for the baselines.

\section{Experiments}
\label{sec:experiments}

\subsection{Experimental Setup}
\label{sec:setup}

\paragraph{Models and training.}
We use two backbones, Llama3.1-8B~\citep{grattafiori2024llama3} (native 128k context) and Qwen3.5-4B (native 256k context), both in their base versions without instruction tuning, so that instruction following comes entirely from the training data and comparisons between datasets are not confounded by the models' original alignment behavior. All training uses the same configuration: full-parameter fine-tuning with the ms-swift framework~\citep{zhao2025swift} on 8 H200 GPUs, learning rate $2\times10^{-5}$ with cosine decay to $2\times10^{-6}$, warmup ratio 0.05, weight decay 0.1, 8 packed sequences per step, and two epochs. Each dataset contributes 2{,}048 training samples, matching the size of the released training sets of SEALONG and LongFaith. For each dataset we sample and train once with each of the random seeds 42, 3407, and 2026, and report the mean and standard deviation over the three seeds.

\paragraph{Baselines.}
We compare with the seven mainstream long-context instruction datasets of Section~\ref{sec:related}: LongAlpaca, LongAlign, LongReward, LongMIT, LongMagpie, SEALONG, and LongFaith. SEALONG releases only preference pairs, from which we take the chosen responses as supervision targets. All seven baselines are normalized into (document, question, answer) triples, with ``document + blank line + question'' as the input and the answer as the output, without any extra instruction; the two mixtures use the same format, and the later turns of \lfn{} give only the questions.

\paragraph{Benchmarks.}
We evaluate on six long-context benchmarks: LongBench~\citep{bai2024longbench}, LongBench v2~\citep{bai2025longbenchv2}, LooGLE~\citep{li2024loogle}, MRCR~\citep{vodrahalli2024michelangelo}, HELMET~\citep{yen2025helmet}, and RULER~\citep{hsieh2024ruler}. Because evaluation has to be repeated across several seeds, for the larger benchmarks we evaluate only discriminative and representative subtasks, applied identically to all models: for HELMET, the four categories retrieval-augmented generation, in-context learning, re-ranking, and long-document QA; for LooGLE, the two QA subtasks; for MRCR, the score pooled over the five bins up to 128k, with the longer 256k bin reported separately in Appendix~\ref{app:mrcr256}. All inference uses greedy decoding. We additionally check the retention of short-context ability on nine general short-context benchmarks in Appendix~\ref{app:short}.

\subsection{Main Results}
\label{sec:main-results}

\begin{table}[t]
  \centering
  \small
  \setlength{\tabcolsep}{4pt}
  \setlength{\aboverulesep}{0pt}
  \setlength{\belowrulesep}{0pt}
  \renewcommand{\arraystretch}{1.12}
  \begin{tabular}{l!{\color{black!40}\vrule}cccccc!{\color{black!40}\vrule}c}
    \toprule
    \textbf{Training data} & \textbf{LongBench} & \textbf{LongBench v2} & \textbf{LooGLE} & \textbf{MRCR} & \textbf{HELMET} & \textbf{RULER} & \textbf{Avg.} \\
    \midrule
    \multicolumn{8}{>{\columncolor{groupbg}}l}{\textit{Llama3.1-8B}} \\
    \midrule
    + LongAlign & 35.37\std{0.53} & 25.23\std{1.66} & 61.67\std{0.26} & 1.54\std{1.29} & 42.15\std{1.39} & 81.99\std{1.45} & 41.32\std{0.51} \\
    + LongAlpaca & 30.78\std{0.27} & 26.17\std{3.92} & 62.08\std{0.73} & 0.02\std{0.03} & 42.91\std{1.47} & 80.33\std{1.76} & 40.38\std{0.52} \\
    + LongReward & 35.28\std{0.73} & 26.53\std{0.70} & 61.78\std{0.30} & 4.60\std{2.57} & 49.07\std{1.51} & 83.81\std{0.71} & 43.51\std{0.82} \\
    + SEALONG & 23.02\std{0.61} & 19.23\std{1.45} & 43.35\std{3.17} & 0.85\std{0.73} & 44.32\std{3.51} & 83.31\std{0.75} & 35.68\std{0.44} \\
    + LongMIT & 33.69\std{0.91} & 24.27\std{0.67} & 63.01\std{0.56} & 11.98\std{3.15} & 46.71\std{0.93} & 86.40\std{1.09} & 44.34\std{0.46} \\
    + LongMagpie & 35.92\std{0.20} & 29.87\std{2.04} & 60.72\std{2.16} & 12.55\std{6.89} & 50.70\std{2.00} & 85.77\std{2.43} & 45.92\std{1.44} \\
    + LongFaith & 23.67\std{0.42} & 20.77\std{3.31} & 51.73\std{0.73} & 0.00\std{0.00} & 47.05\std{3.54} & 75.94\std{3.06} & 36.53\std{1.78} \\
    \midrule
    \rowcolor{oursbg}+ \textbf{\lfone{}} & 41.56\std{0.09} & 28.20\std{1.78} & 64.96\std{0.28} & \textbf{19.29}\std{2.18} & 53.35\std{0.85} & 85.84\std{0.43} & 48.87\std{0.64} \\
    \rowcolor{oursbg}+ \textbf{\lfn{}} & \textbf{42.27}\std{1.10} & \textbf{33.00}\std{1.20} & \textbf{66.58}\std{0.51} & 15.92\std{1.84} & \textbf{56.71}\std{0.39} & \textbf{89.28}\std{0.78} & \textbf{50.63}\std{0.37} \\
    \midrule
    \multicolumn{8}{>{\columncolor{groupbg}}l}{\textit{Qwen3.5-4B}} \\
    \midrule
    + LongAlign & 35.74\std{0.33} & 30.27\std{0.12} & 65.82\std{1.53} & 30.07\std{3.48} & 51.15\std{2.73} & 88.93\std{1.46} & 50.33\std{0.60} \\
    + LongAlpaca & 40.61\std{1.30} & 33.80\std{1.51} & 69.29\std{0.99} & 18.90\std{3.63} & 57.11\std{0.37} & 89.74\std{0.29} & 51.57\std{0.86} \\
    + LongReward & 38.62\std{1.73} & 33.20\std{1.00} & 66.45\std{0.12} & 32.38\std{0.77} & 55.59\std{0.21} & 91.42\std{0.33} & 52.94\std{0.34} \\
    + SEALONG & 26.50\std{0.21} & 28.27\std{0.23} & 49.72\std{0.94} & 2.48\std{2.89} & 52.38\std{1.27} & 86.45\std{0.98} & 40.97\std{0.80} \\
    + LongMIT & 40.58\std{1.03} & 30.27\std{0.31} & 65.32\std{0.21} & 30.32\std{0.83} & 55.23\std{0.87} & 88.82\std{0.96} & 51.76\std{0.07} \\
    + LongMagpie & 45.42\std{0.10} & 34.33\std{1.33} & 65.90\std{0.88} & 32.75\std{0.99} & 57.63\std{0.87} & \textbf{92.41}\std{0.18} & 54.74\std{0.36} \\
    + LongFaith & 28.75\std{0.26} & 36.07\std{1.53} & 64.95\std{0.43} & 22.00\std{5.24} & 60.19\std{0.23} & 85.29\std{0.37} & 49.54\std{0.84} \\
    \midrule
    \rowcolor{oursbg}+ \textbf{\lfone{}} & 46.34\std{0.35} & 37.00\std{0.20} & \textbf{71.39}\std{0.56} & 35.18\std{1.25} & 60.61\std{0.61} & 90.87\std{0.56} & 56.90\std{0.10} \\
    \rowcolor{oursbg}+ \textbf{\lfn{}} & \textbf{47.69}\std{0.19} & \textbf{38.93}\std{0.81} & 71.28\std{0.77} & \textbf{37.08}\std{0.43} & \textbf{60.95}\std{0.88} & 91.30\std{0.74} & \textbf{57.87}\std{0.27} \\
    \bottomrule
  \end{tabular}
  \caption{Six-benchmark results on the two backbones (mean and std over three seeds); shaded rows are ours, and bold marks the best per backbone.}
  \label{tab:main}
\end{table}

Table~\ref{tab:main} gives the six-benchmark results on the two backbones, visualized in Figure~\ref{fig:main} on the first page, with the subtask-level readings of each benchmark in Appendix~\ref{app:breakdown}. The six-benchmark averages of \lfone{} are 48.87 on Llama3.1-8B and 56.90 on Qwen3.5-4B, 2.95 and 2.16 points above the strongest baseline, LongMagpie, and it exceeds that baseline on five of six benchmarks on both backbones; its sample length is unremarkable among the training sets (Appendix~\ref{app:tokens}). The gains are largest on tasks that require understanding and integrating long-context content: $+4.2$ and $+5.5$ points on LooGLE and $+2.7$ and $+3.0$ on HELMET, whereas on the retrieval-oriented RULER \lfone{} and LongMagpie differ by under 1.6 points. The gain of \lfone{} thus lies mainly in integrating and reasoning over long-context content rather than in retrieval.

\lfn{} reaches 50.63 and 57.87 on the two backbones, 4.71 and 3.13 points above the strongest baseline and a further 1.76 and 0.97 points above \lfone{}. Compared with \lfone{}, \lfn{} raises document utilization from 0.102 to 0.519 and also packs more QA pairs into one sample; yet at a comparable total number of QA pairs, i.e., scaling \lfone{} to 8{,}192 samples, the Qwen3.5-4B six-benchmark average changes by only $+0.12$ (Appendix~\ref{app:scale}). Letting one training sample cover more of the document thus helps bring downstream gains.

\subsection{Ablations}
\label{sec:ablation}

\begin{table}[t]
  \centering
  \small
  \setlength{\tabcolsep}{4pt}
  \setlength{\aboverulesep}{0pt}
  \setlength{\belowrulesep}{0pt}
  \renewcommand{\arraystretch}{1.12}
  \resizebox{\textwidth}{!}{%
  \begin{tabular}{l!{\color{black!40}\vrule}c!{\color{black!40}\vrule}ccc}
    \toprule
    \textbf{Item} & \textbf{\lfone{} (control)} & \textbf{Document source} & \textbf{Synthesis model} & \textbf{Distribution mixture} \\
    \midrule
    \rowcolor{groupbg}Document pool & Mixed pool & \cellcolor{changedbg}\textit{LongMagpie only} & Mixed pool & Mixed pool \\
    \rowcolor{groupbg}Synthesis model & DeepSeek-V4.1-Flash & DeepSeek-V4.1-Flash & \cellcolor{changedbg}\textit{Qwen3.8-27B} & DeepSeek-V4.1-Flash \\
    \rowcolor{groupbg}Breadth $B$ (mean / median) & 12.2 / 10 & 11.7 / 10 & 11.2 / 9 & \cellcolor{changedbg}\textit{4.41 / 5} \\
    \rowcolor{groupbg}Depth $P$ & 2.89 & 3.12 & 2.74 & \cellcolor{changedbg}\textit{1.95} \\
    \rowcolor{groupbg}Single-step rate & 0.154 & 0.092 & 0.210 & \cellcolor{changedbg}\textit{0.462} \\
    \midrule
    LongBench & 46.34\std{0.35} & 45.67\std{0.65} & 48.53\std{0.52} & 45.57\std{0.45} \\
    LongBench v2 & 37.00\std{0.20} & 38.00\std{1.71} & 36.13\std{2.16} & 35.53\std{0.58} \\
    LooGLE & 71.39\std{0.56} & 70.80\std{0.43} & 70.47\std{0.37} & 69.33\std{0.14} \\
    MRCR & 35.18\std{1.25} & 33.61\std{0.61} & 36.42\std{1.64} & 34.33\std{1.14} \\
    HELMET & 60.61\std{0.61} & 59.69\std{0.06} & 60.15\std{0.73} & 59.26\std{0.46} \\
    RULER & 90.87\std{0.56} & 90.37\std{0.78} & 91.54\std{0.37} & 90.43\std{0.60} \\
    \midrule
    \rowcolor{oursbg}\textbf{Avg.} & 56.90\std{0.10} & 56.36\std{0.26} & 57.21\std{0.62} & 55.74\std{0.11} \\
    \rowcolor{oursbg}$\Delta$ \textbf{Avg.} & -- & $-0.54$ & $+0.31$ & $-1.16$ \\
    \bottomrule
  \end{tabular}}
  \caption{Settings and results of the three ablation groups. Each group changes one setting (italic) and keeps the rest identical to the control; the first two groups do not change the distribution mixture, so breadth and depth fluctuate slightly with their candidate pools. The seven baselines pooled have breadth 7.18 / 5 (mean / median), depth 1.798, and single-step rate 0.434.}
  \label{tab:ablation}
\end{table}

All ablations are run with \lfone{} on Qwen3.5-4B. We design three groups, studying the effect of the document source, the synthesis model, and the distribution mixture (Table~\ref{tab:ablation}).

\paragraph{Document source.}
Replacing the mixed document pool with one drawn from LongMagpie alone lowers the average by 0.54 points, but it is still 1.62 points above the strongest baseline, LongMagpie, so the gain of our method does not come from mixing document pools.

\paragraph{Synthesis model.}
Replacing the synthesis model DeepSeek-V4.1-Flash with Qwen3.8-27B gives an average of 57.21, 0.31 points above the control. Downstream results do not drop with a smaller model, so the method does not depend on one particular strong synthesis model.

\paragraph{Distribution mixture.}
With the gate and the candidate QA pairs identical to the control, mixing breadth and depth down to the baseline level lowers the six-benchmark average from 56.90 to 55.74, a drop of 1.16 points, with all six benchmarks below the control. Read the other way, raising breadth and depth is worth 1.16 points on its own, with everything else about the data held fixed: the structural distribution is a lever on downstream performance, not merely a description of the data.

\subsection{Case Study}
\label{sec:case}

\begin{figure}[t]
  \centering
  \includegraphics[width=\textwidth]{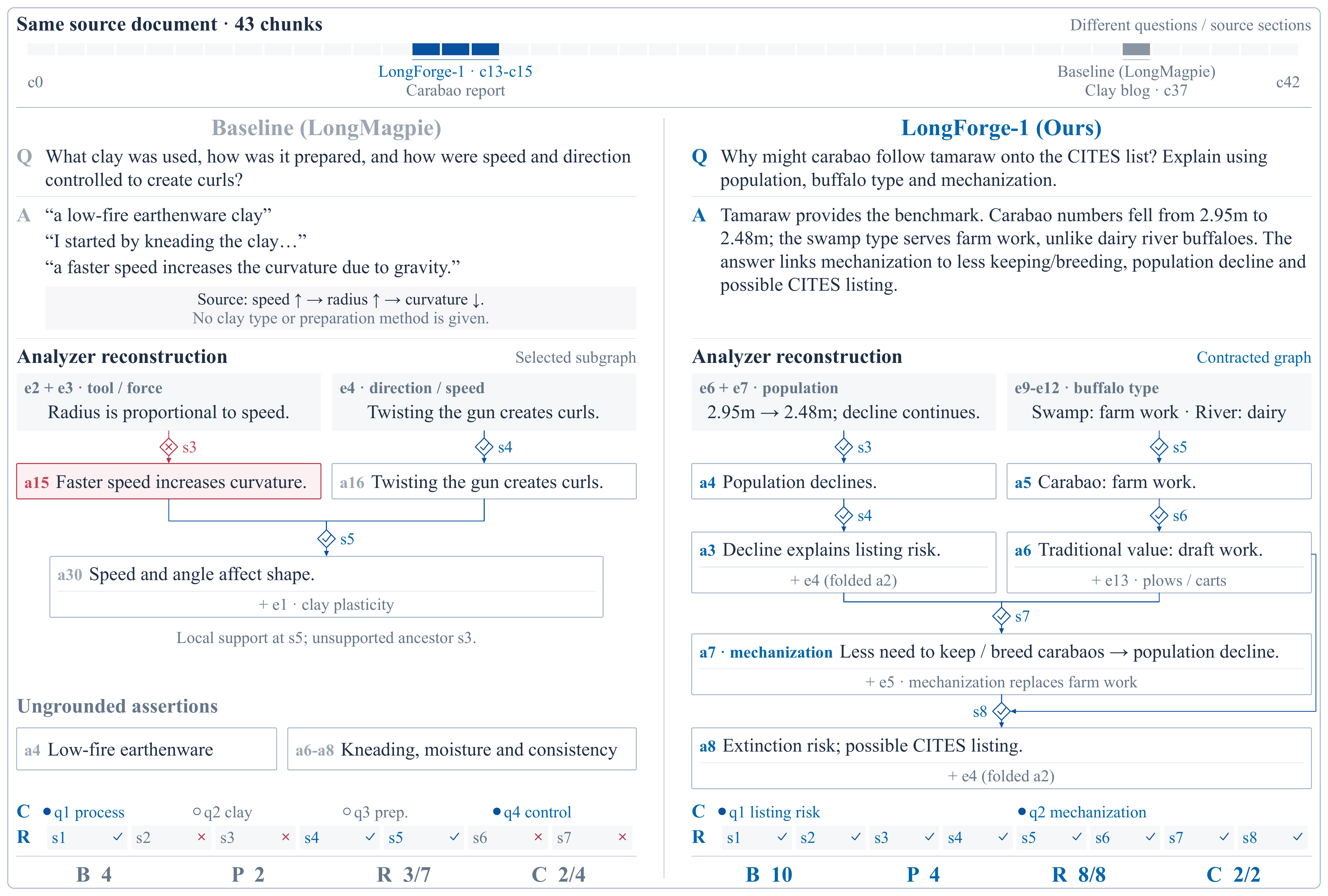}
  \caption{A LongMagpie (left) and a \lfone{} (right) QA pair on the same source document. Top: the positions in the document of the evidence each pair uses. Middle: the question, the answer, and the reasoning graph the analyzer reconstructs from them, where diamonds are reasoning steps and red marks an unsupported one. Bottom: information-point coverage, the step judgments, and the readings of the four dimensions.}
  \label{fig:case1}
\end{figure}

To show what the four dimensions mean on a concrete sample, we contrast one baseline and one \lfone{} QA pair on the same source document (Figure~\ref{fig:case1}); an example of how \lfn{} covers a long document in a complementary way is in Appendix~\ref{app:case-n}.

The two QA pairs differ on different dimensions. The baseline answer addresses every part of the question in wording, but some whole sentences find no support in the document and become isolated assertions, and one step distorts a relation in the source and is judged unsupported, so the reasoning built on it also loses its ground; both correctness and completeness are therefore low. The \lfone{} answer starts from evidence in several places and organizes two reasoning branches that merge at the terminal assertion, with every step supported and every information point answered from evidence, and both breadth and depth above the baseline. The contrast shows that the analyzer can locate the quality difference between two QA pairs at specific evidence spans, assertions, and reasoning steps, rather than only returning a higher or lower score.

\section{Conclusion}
\label{sec:conclusion}

We argue that long-context data quality should be represented as a traceable structure rather than a scalar. Defined on the alignment among document, question, and answer, quality becomes measurable: how much of the document a sample uses, how much reasoning it writes down, and which step or requirement fails. This measurement exposes the narrow evidence use and shallow reasoning of existing data, lets the synthesizer re-answer from located failures, and makes the training distribution settable; raising breadth and depth improves downstream performance on its own, so the structure of the training distribution is a design choice a data pipeline can make deliberately rather than a property it inherits. We hope the reasoning graph and its four dimensions become a common measure for long-context data.

\section*{Limitations}

Measurement is costly: the analyzer scans every chunk for evidence, so the number of calls grows linearly with document length and dominates the synthesis cost (Appendix~\ref{app:synth-stats}). This keeps breadth and utilization free of retrieval misses, and a smaller extraction model may lower the cost. In addition, we compare only three distributions on the synthesized data, namely the baseline-level mixture, \lfone{}, and \lfn{}; they show that raising breadth and depth helps, but not how far to raise them or how much each contributes, which we leave to a more systematic mixture design.

\section*{Ethical Considerations}

All seven baseline datasets are publicly released, under CC BY-NC 4.0 (LongAlpaca), Apache-2.0 (LongAlign, LongReward, LongMIT, LongMagpie), or MIT (SEALONG, LongFaith), and our use of them for training, measurement, and synthesis is limited to academic research in accordance with these licenses. The synthesized QA pairs use their documents as sources and involve no newly collected personal information. Being generated by language models, they may inherit biases or inappropriate content from the source documents; since the gate checks only whether answers follow from the documents and address the questions, not whether the documents themselves are accurate or appropriate, the data is intended for research use only.

\bibliography{references}

\newpage
\appendix

\FloatBarrier
\section{Synthesis Details}
\label{app:synthesis}

\subsection{Prompts}
\label{app:prompts}

The original prompts of the synthesizer are given below; \texttt{\{...\}} denotes a placeholder.

\begin{promptbox}{Question generation}
[system]
You are an expert at generating questions over long documents.

[user]
Read the document below and write {n} questions about it.
Requirements:
- answering the question must require combining evidence from several separate places in the document;
- answering the question must require multiple steps of reasoning;
- the {n} questions should each rely on as different a set of evidence as possible;
- write every question in the same language as the document.
Return JSON, one string per question:
{"questions": ["...", "..."]}

=== DOCUMENT ===
{document}
=== END DOCUMENT ===
\end{promptbox}

\begin{promptbox}{First answer}
[system]
Answer the question based on the content of the document.

[user]
{document}

{question}
\end{promptbox}

\begin{promptbox}{Re-answer}
[system]
Answer the question based on the content of the document.

[user]
{document}

{question}

=== PREVIOUS ANSWER ===
{previous_answer}
=== END PREVIOUS ANSWER ===
=== PROBLEMS FOUND ===
{feedback_block}
=== END PROBLEMS FOUND ===
The previous answer was checked against the document and did not hold up, for the reasons listed above. Write a new answer that fixes them.
\end{promptbox}

\begin{promptbox}{Feedback lines in \texttt{\{feedback\_block\}} (at most 6 per re-answer)}
- [unanswered] The question asks for: {need}
- [unsupported] This claim does not follow from the document: {claim} (checker's note: {reason})
\end{promptbox}

\subsection{Synthesis Statistics}
\label{app:synth-stats}

This section reports statistics of the entire synthesis pool (4{,}096 documents with five questions each, 20{,}480 QA pairs in total).

\begin{table}[!ht]
  \centering
  \small
  \begin{tabular}{l!{\color{black!40}\vrule}cc}
    \toprule
    Round & Passed in round & Cumulative pass rate \\
    \midrule
    1 (first answer) & 17{,}912 & 87.46\% \\
    2 & 1{,}943 & 96.95\% \\
    3 & 346 & 98.64\% \\
    4 & 117 & 99.21\% \\
    Finally failed & 162 & -- \\
    \bottomrule
  \end{tabular}
  \caption{Gate pass statistics of each answering round during synthesis.}
  \label{tab:rounds}
\end{table}

\paragraph{Question generation.}
All 4{,}096 documents receive five questions; 4 documents are retried once because their output could not be parsed.

\paragraph{Answering and gating.}
Table~\ref{tab:rounds} lists the number of QA pairs that pass the gate in each answering round; round 1 is the first answer, and rounds 2 to 4 are re-answers. Each QA pair takes 1.17 attempts on average. Of the 2{,}568 QA pairs that fail the first attempt, 2{,}406 (93.7\%) are repaired by re-answering; the failure reasons are dominated by correctness (2{,}196 fail only correctness, 292 fail only completeness, 21 fail both, and 59 are degenerate or empty answers). Of the 162 finally failed QA pairs, 157 fail only correctness. Among the retained answers, 94.1\% have at least 6 reasoning steps and fall into the most lenient tier of the correctness threshold, so this tier determines the overall pass rate of the gate.

\paragraph{Comparison with baselines.}
If the same gate is applied to the seven baselines, 55.5\% of the samples pass both correctness and completeness. The first-answer pass rate of the synthesizer is 87.5\%, 32 percentage points higher than the baselines, reflecting the quality of generation itself; re-answering further raises the pass rate to 99.2\%, reflecting the repair effect of re-answering with failure reasons.

\begin{table}[!ht]
  \centering
  \small
  \setlength{\tabcolsep}{6pt}
  \begin{tabular}{ll!{\color{black!40}\vrule}ccc}
    \toprule
    Component & Stage & Per doc. & Per QA & Share \\
    \midrule
    \multirow{3}{*}{Synthesizer} & Question generation & 1.0 & 0.20 & 0.15\% \\
    & Answering (incl.\ re-answers) & 5.8 & 1.17 & 0.86\% \\
    & \textit{Subtotal} & \textit{6.8} & \textit{1.37} & \textit{1.01\%} \\
    \midrule
    \multirow{4}{*}{Analyzer} & Evidence extraction & 518.3 & 103.7 & 76.49\% \\
    & Graph assembly & 25.5 & 5.1 & 3.76\% \\
    & Step judgment & 126.9 & 25.4 & 18.73\% \\
    & \textit{Subtotal} & \textit{670.8} & \textit{134.2} & \textit{98.99\%} \\
    \midrule
    \textbf{Total} & & \textbf{677.6} & \textbf{135.5} & 100\% \\
    \bottomrule
  \end{tabular}
  \caption{Model-call cost of synthesis and analysis. ``Per QA'' counts all 20{,}480 generated QA pairs, whereas the per-QA cost in the text is normalized by the pairs that pass the gate.}
  \label{tab:cost}
\end{table}

\paragraph{Call cost.}
Table~\ref{tab:cost} measures cost by the number of model calls, counting the synthesizer and the analyzer separately. Normalized by the QA pairs that finally pass the gate, each consumes 136.6 calls on average, 1.38 by the synthesizer and 135.2 by the analyzer; the entire synthesis pool takes about 2.78 million calls. Measurement, not generation, thus dominates the cost, and evidence extraction alone accounts for three quarters of it, a direct consequence of traversing every chunk rather than prefiltering by retrieval.

\FloatBarrier
\section{Token Statistics of the Training Data}
\label{app:tokens}

\begin{table}[!ht]
  \centering
  \small
  \setlength{\tabcolsep}{6pt}
  \begin{tabular}{l!{\color{black!40}\vrule}ccc!{\color{black!40}\vrule}ccc}
    \toprule
    & \multicolumn{3}{c!{\color{black!40}\vrule}}{Qwen tokenizer} & \multicolumn{3}{c}{Llama tokenizer} \\
    \cmidrule(lr){2-4} \cmidrule(lr){5-7}
    Data & Total & Input & Output & Total & Input & Output \\
    \midrule
    LongAlign & 16254 & 16072 & 182 & 16974 & 16773 & 201 \\
    LongAlpaca & 12828 & 12559 & 270 & 12409 & 12144 & 265 \\
    LongReward & 22136 & 21858 & 278 & 23555 & 23214 & 341 \\
    SEALONG & 19421 & 19157 & 264 & 18465 & 18212 & 253 \\
    LongMIT & \textbf{69699} & \textbf{69554} & 145 & \textbf{78131} & \textbf{77957} & 174 \\
    LongMagpie & 6250 & 5496 & 754 & 6051 & 5322 & 729 \\
    LongFaith & 2871 & 2643 & 228 & 2726 & 2508 & 218 \\
    \textbf{\lfone{}} & 18248 & 17726 & 522 & 18803 & 18268 & 535 \\
    \textbf{\lfn{}} & 23190 & 19805 & \textbf{3385} & 24026 & 20518 & \textbf{3508} \\
    \bottomrule
  \end{tabular}
  \caption{Token lengths of each training set.}
  \label{tab:tokens}
\end{table}

Table~\ref{tab:tokens} reports the mean token length of each training set, averaged over the three seeds with 2{,}048 samples each. The input consists of the document and the question, and the output is the answer; for \lfn{}, the input contains the document and the questions of all turns, and the output is the sum of the answers of all turns.

The two tokenizers give largely the same ranking. Neither the input nor the output of \lfone{} is the longest: its total and input lengths are both below those of LongMIT and LongReward, and its output is below that of LongMagpie. Yet the downstream performance of \lfone{} exceeds all baselines, so the downstream gain cannot be attributed to longer samples.

The input of \lfn{} is about 1.12 times that of \lfone{}, while its output reaches 6.5 times that of \lfone{} because each sample contains 4.96 turns of answers on average. This increment comes from the larger number of QA pairs, but the number of QA pairs alone does not explain the improvement of \lfn{}: \lfone{} with 8{,}192 samples contains a comparable number of QA pairs to \lfn{} (about 10{,}000 per seed), yet its six-benchmark average is still lower (57.02 vs.\ 57.87, Appendix~\ref{app:scale}). The gain of \lfn{} thus more likely comes from the increase in breadth, i.e., a sample covering more of the document.

\FloatBarrier
\section{Additional Statistical Analyses}
\label{app:analysis}

\subsection{Joint Distribution of Breadth and Depth per Baseline}
\label{app:joint}

\begin{figure}[!ht]
  \centering
  \includegraphics[width=0.84\textwidth]{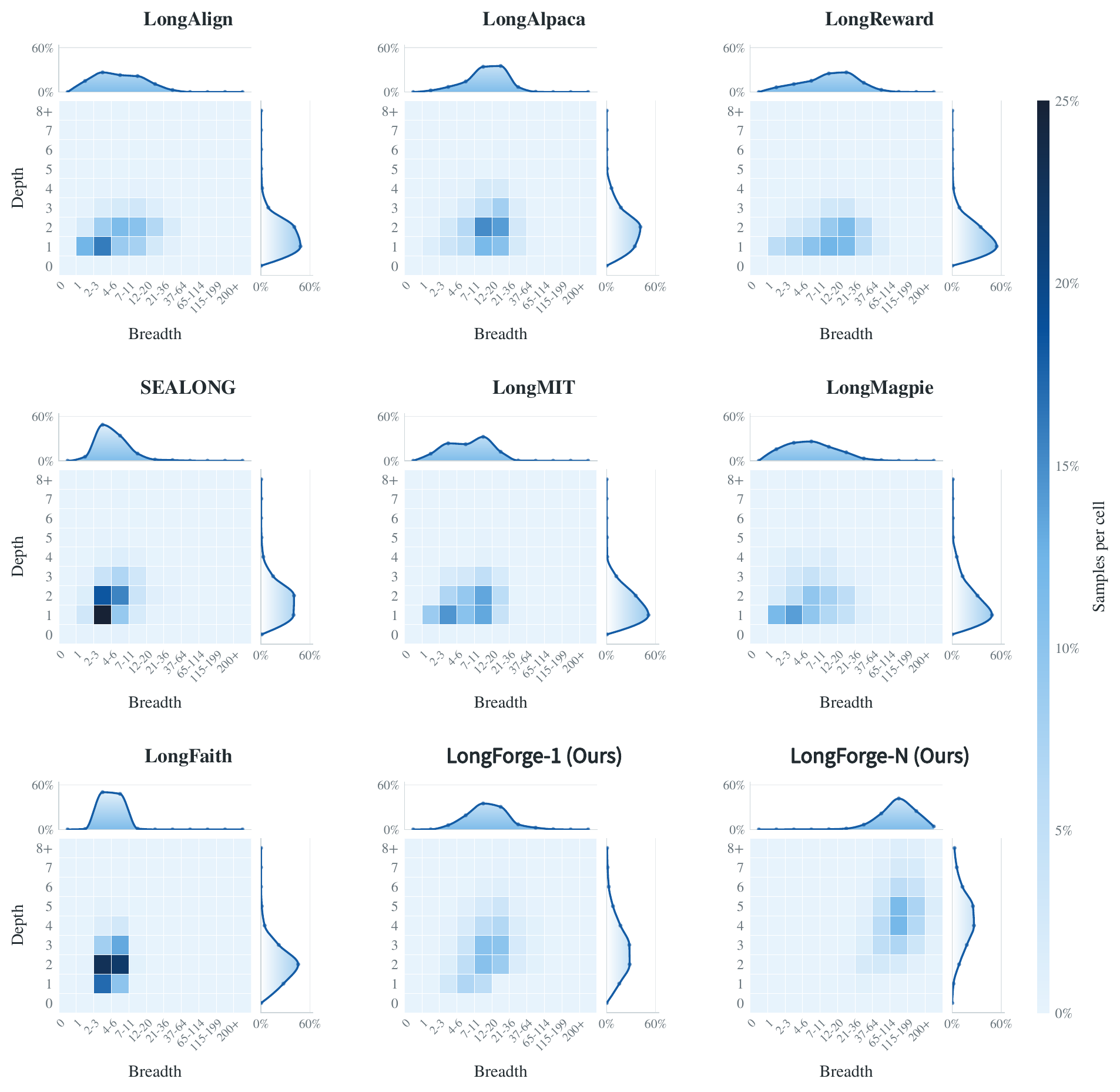}
  \caption{Joint distribution of breadth and depth with the seven baselines shown separately; the last two panels are our two mixtures.}
  \label{fig:by-baseline}
\end{figure}

Figure~\ref{fig:joint} pools the seven baselines into one overall distribution. This section separates them and places them alongside the two mixtures in nine panels (Figure~\ref{fig:by-baseline}); Table~\ref{tab:by-baseline} gives summary readings for each panel.

\begin{table}[!ht]
  \centering
  \small
  \resizebox{\textwidth}{!}{%
  \begin{tabular}{lcccccc}
    \toprule
    Dataset & Samples & Breadth (median / mean) & Depth (median / mean) & Single-step rate & Wide-and-deep & Modal cell (depth / breadth) \\
    \midrule
    LongAlign & 994 & 4.5 / 6.27 & 2 / 1.649 & 48.3\% & 28.6\% & 1 / 2--3 \\
    LongAlpaca & 963 & \textbf{10} / 11.17 & 2 / 1.977 & 34.5\% & \textbf{55.9\%} & 2 / 7--11 \\
    LongReward & 986 & \textbf{10} / \textbf{12.43} & 1 / 1.595 & 54.4\% & 38.7\% & 1 / 7--11 \\
    SEALONG & 1000 & 3 / 4.08 & 2 / 1.835 & 39.8\% & 15.0\% & 1 / 2--3 \\
    LongMIT & 1000 & 6 / 6.44 & 1 / 1.621 & 51.2\% & 31.4\% & 1 / 2--3 \\
    LongMagpie & 891 & 5 / 6.51 & 2 / 1.860 & 48.6\% & 25.6\% & 1 / 2--3 \\
    LongFaith & 1000 & 3 / 3.51 & 2 / \textbf{2.056} & \textbf{27.4\%} & 5.1\% & 2 / 2--3 \\
    \midrule
    \textit{Seven pooled} & \textit{6834} & \textit{5 / 7.18} & \textit{2 / 1.798} & \textit{43.4\%} & \textit{28.5\%} & \textit{1 / 2--3} \\
    \bottomrule
  \end{tabular}}
  \caption{Summary readings of breadth and depth per dataset; the last row is the pooled total. Wide-and-deep means breadth at least 6 and depth at least 2; the modal cell is the cell with the largest share of samples in Figure~\ref{fig:by-baseline}, given by its depth and breadth interval.}
  \label{tab:by-baseline}
\end{table}

\paragraph{All seven datasets concentrate in the lower-left corner, but at different positions.}
The median breadth ranges from 3 (SEALONG, LongFaith) to 10 (LongAlpaca, LongReward), and the wide-and-deep share from 5.1\% to 55.9\%. Four of the seven have the modal cell ``depth 1, breadth 2--3,'' i.e., single-step with only two or three evidence spans; the modal cell of the pooled total is also this cell (14.0\%), showing that this is the most common form of existing long-context data rather than an artifact of pooling.

\paragraph{The two dimensions can be high or low independently.}
LongReward ties with LongAlpaca for the highest median breadth (10), yet its single-step rate is the highest of the seven (54.4\%); LongFaith is the opposite, with the lowest single-step rate (27.4\%) but a median breadth of only 3. One uses much evidence but reasons shallowly, and the other writes deep reasoning with little evidence, a direct manifestation of the two dimensions reading different positions of the reasoning graph.

\paragraph{The near-orthogonality also holds within each dataset.}
The Pearson correlations within the seven datasets range from 0.025 (LongMagpie) to 0.258 (LongMIT), and the pooled value after centering by dataset is 0.111, on the same order as the overall 0.070. The low overall correlation thus reflects genuine near-orthogonality of breadth and depth rather than an artifact of pooling.

\subsection{Correlations among the Four Dimensions}
\label{app:correlation}

The four dimensions read different positions of the reasoning graph; this section tests on data whether they indeed cannot substitute for one another (Table~\ref{tab:corr}).

\begin{table}[!ht]
  \centering
  \small
  \setlength{\tabcolsep}{8pt}
  \begin{tabular}{l!{\color{black!40}\vrule}cccc}
    \toprule
    & Breadth & Depth & Correctness & Completeness \\
    \midrule
    \textbf{Breadth} & -- & 0.189 & $-0.134$ & 0.109 \\
    \textbf{Depth} & 0.070 & -- & $-0.253$ & 0.010 \\
    \textbf{Correctness} & $-0.025$ & $-0.166$ & -- & 0.032 \\
    \textbf{Completeness} & 0.100 & $-0.007$ & 0.035 & -- \\
    \bottomrule
  \end{tabular}
  \caption{Correlations among the four dimensions over the 6{,}834 baseline samples (upper triangle: Spearman; lower triangle: Pearson).}
  \label{tab:corr}
\end{table}

The absolute values of all six correlations are at most 0.253, i.e., any two dimensions share at most 6.4\% of their variance, and no dimension can be approximated by another. The strongest pair is the negative rank correlation between depth and correctness ($-0.253$), i.e., the finding in Section~\ref{sec:measurement} that the longer the reasoning chain, the more likely one step is unsupported.

\subsection{Definitions of the Failure Axes}
\label{app:failure}

Table~\ref{tab:axes} defines the seven failure axes used in Section~\ref{sec:measurement}. Each axis is aggregated within a dataset from the per-sample quantities of Section~\ref{sec:analyzer} and oriented so that higher is worse; within each axis, we apply z-standardization with the mean $\mu$ and the population standard deviation $\delta$ over the seven datasets, $z = (v - \mu) / \delta$, where $v$ is the value of a dataset on the axis, and the axis with the highest $z$ is the dataset's dominant defect.

\begin{table}[!ht]
  \centering
  \footnotesize
  \setlength{\tabcolsep}{4pt}
  \begin{tabular}{l l p{0.34\textwidth} p{0.26\textwidth}}
    \toprule
    Dimension & Axis & Computation (aggregated within a dataset) & Meaning \\
    \midrule
    Breadth & Narrow & Share of samples with breadth at most 2 & The answer uses at most two evidence spans \\
    Breadth & Local & Sample mean of $1 -$ evidence spread$/(M - 1)$; the evidence spread is the index difference between the first and last chunks that yield evidence, and the denominator is 1 when $M = 1$ & Evidence is clustered in a small part of the document \\
    Breadth & Doc-starved & Sample mean of $1 -$ document utilization; document utilization is the number of chunks that yield evidence$/M$ & Most of the document is never used \\
    Depth & Shallow & Share of samples with depth 1 & No explicit intermediate reasoning is written down \\
    Correctness & Ungrounded & Share of isolated assertions among all assertions of a sample, averaged over samples & Some whole answer sentences connect to no evidence \\
    Correctness & Unsupported & Sample mean of $1 -$ correctness & An assertion connects to evidence, but the step does not derive it \\
    Completeness & Incomplete & Sample mean of $1 -$ completeness & Some information points of the question are unanswered \\
    \bottomrule
  \end{tabular}
  \caption{Definitions of the seven failure axes. $M$ is the number of chunks in the document; the axis names are the column names in Figure~\ref{fig:failure}.}
  \label{tab:axes}
\end{table}

\subsection{Case Study of \lfn{}}
\label{app:case-n}

\begin{figure}[!ht]
  \centering
  \includegraphics[width=0.58\textwidth]{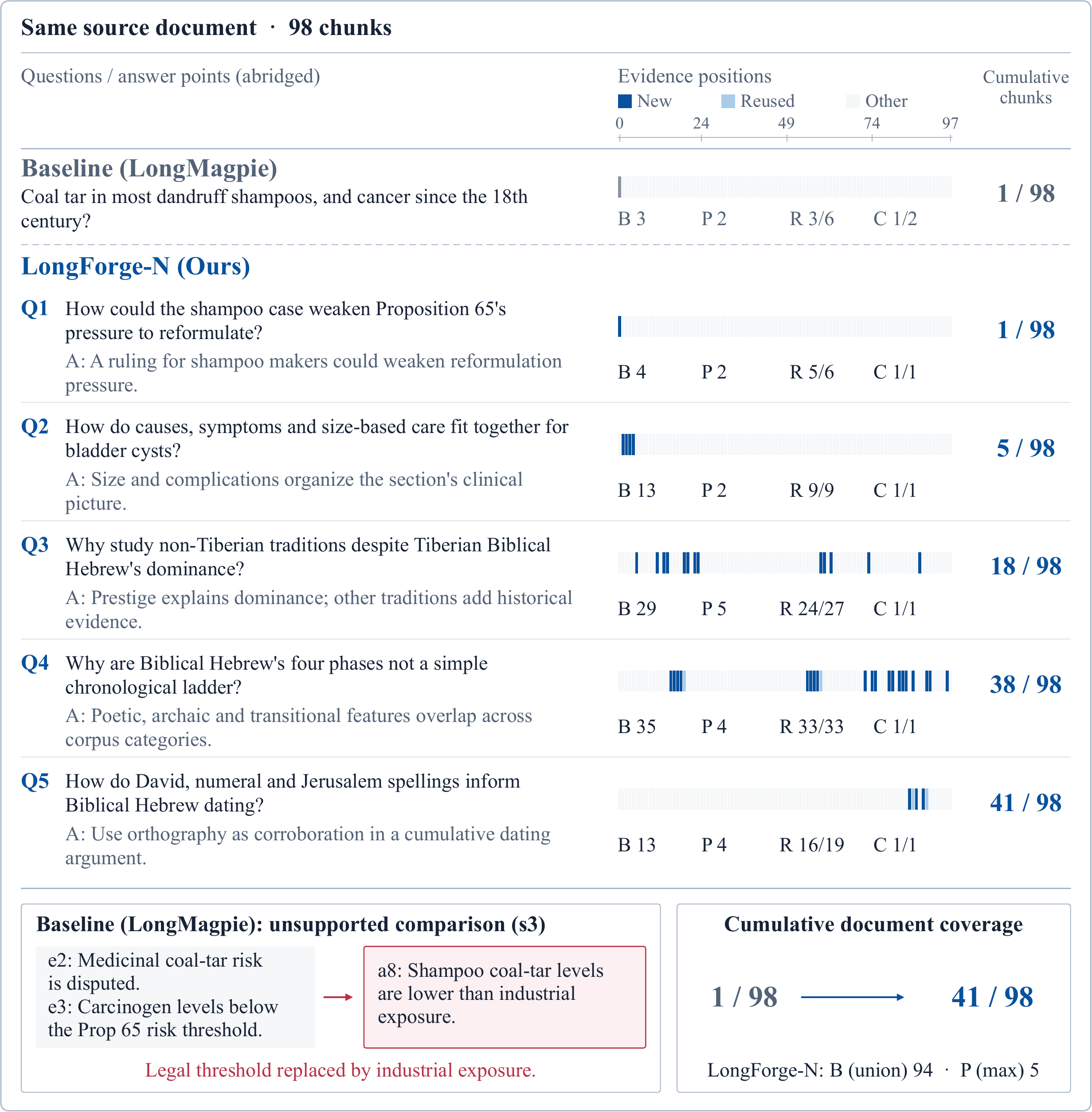}
  \caption{One QA pair from LongMagpie contrasted with five turns of QA from \lfn{} on the same source document. On the right of each row are the positions in the document of the evidence used by that QA pair: dark chunks are new (unused in earlier turns), light chunks are reused, and the rightmost column gives the cumulative number of covered chunks. Bottom left: a local reasoning graph of an unsupported step in the baseline. Bottom right: the cumulative document coverage of the five turns.}
  \label{fig:case-n}
\end{figure}

Figure~\ref{fig:case-n} shows the multi-turn QA that \lfn{} organizes on one document. The baseline QA pair concentrates on a small segment at the beginning of the document; the questions of \lfn{}, in contrast, fall successively on different parts of the document, later questions keep introducing previously unused content with little reuse, and the cumulative coverage expands turn by turn. Each question also requires integrating arguments from several places to answer. This shows that, through a set of complementary questions, content scattered across a long document can be organized into diverse training signals.

The local reasoning graph at the bottom left also shows the diagnostic role of the analyzer: when paraphrasing the source text, the baseline changes the object of comparison and introduces a conclusion unsupported by the source; following the connections between evidence and assertions locates the step where the deviation occurs, so the quality problem of the answer can be explained concretely.

\subsection{Human Audit Details}
\label{app:audit}

\paragraph{Sampling and auditors.}
We draw 350 samples from the seven baseline datasets, stratified by source; after removing 4 degenerate samples for which no reasoning graph could be built, 346 samples remain for auditing. Both auditors are PhD students working on long-context research with experience in long-context annotation; their compensation matches the local internship pay for PhD students, and both were informed of the audit content and the data use before the audit and agreed to participate.

\paragraph{Audit procedure and criteria.}
The two auditors independently check all samples. For each dimension, an auditor first derives the value on the reasoning graph on their own, then compares it with the analyzer and rates it as consistent, minor deviation, or major deviation according to Table~\ref{tab:criteria}. The main text reports the results of the stricter auditor, who rates more major deviations.

\begin{table}[!ht]
  \centering
  \scriptsize
  \setlength{\tabcolsep}{4pt}
  \begin{tabular}{p{0.11\textwidth} p{0.15\textwidth} p{0.18\textwidth} p{0.18\textwidth} p{0.25\textwidth}}
    \toprule
    Dimension & Meaning of the analyzer's value & Consistent & Minor deviation & Major deviation \\
    \midrule
    Breadth & Number of evidence spans counted toward breadth, excluding evidence discarded at assembly & The auditor's evidence count equals the analyzer's, or differs by 1 where that span is a borderline case (quoted range slightly wider or narrower) & Differs by 1--2 without changing the order-of-magnitude judgment of how much of the document the sample uses & Differs by 3 or more; or a counted evidence span is entirely irrelevant to the question; or a discarded evidence span is necessary for the answer \\
    \midrule
    Depth & Number of hops of the longest reasoning path after paraphrase contraction & The longest chain reconstructed by the auditor has the same number of hops as the analyzer's & Differs by 1 hop without crossing the boundary between 1 and 2 & Differs by 2 or more hops; or crosses the boundary between 1 and 2 (1 hop means no intermediate reasoning and 2 hops at least one inference, a qualitative difference); or a step counted as reasoning is a pure paraphrase \\
    \midrule
    Correctness & Share of supported steps among all reasoning steps & The auditor's set of supported steps is identical to the analyzer's & Exactly 1 step is judged differently, without crossing 0.5 & 2 or more steps are judged differently; or crosses 0.5 (majority supported vs.\ majority unsupported); or a step the analyzer judges supported is clearly underivable (evidence unrelated to the assertion, or a proper noun absent from the inputs) \\
    \midrule
    Completeness & Share of answered information points among all information points & The auditor's judgments of whether each information point is answered all match the analyzer's & Some information points are judged differently, without crossing complete vs.\ incomplete & Crosses complete vs.\ incomplete (1.0 vs.\ below 1.0); or the question decomposition itself is wrong (e.g., what should be two information points is decomposed as one) \\
    \bottomrule
  \end{tabular}
  \caption{Audit criteria for the four dimensions.}
  \label{tab:criteria}
\end{table}

\paragraph{Inter-auditor agreement.}
Table~\ref{tab:irr} gives the agreement between the two auditors' independent judgments. Counted as ``usable vs.\ major deviation,'' the agreement on all four dimensions is above 0.90, and Gwet's AC1~\citep{gwet2008ac1}, which is robust to skewed distributions, gives similar readings. When consistent and minor deviation are distinguished, only correctness drops noticeably (0.671), with disagreements concentrated on whether individual reasoning steps are supported.

\begin{table}[!ht]
  \centering
  \small
  \resizebox{\textwidth}{!}{%
  \begin{tabular}{lccccc}
    \toprule
    Dimension & Usable rate (A / B / both) & Binary agreement & Binary AC1 & Three-level agreement & Three-level AC1 \\
    \midrule
    Breadth & 99.7\% / 100\% / 99.7\% & 0.997 & 0.997 & 0.997 & 0.997 \\
    Depth & 97.4\% / 98.6\% / 96.2\% & 0.965 & 0.964 & 0.957 & 0.956 \\
    Correctness & 89.6\% / 95.4\% / 87.9\% & 0.908 & 0.893 & 0.671 & 0.568 \\
    Completeness & 98.0\% / 100\% / 98.0\% & 0.980 & 0.979 & 0.971 & 0.971 \\
    \bottomrule
  \end{tabular}}
  \caption{Agreement between the two auditors. The usable-rate column lists auditor A, auditor B, and the share rated usable by both; the binary scheme distinguishes only usable (consistent or minor deviation) from major deviation, and the three-level scheme distinguishes consistent, minor deviation, and major deviation.}
  \label{tab:irr}
\end{table}

\FloatBarrier
\section{Additional Experiments}
\label{app:extra}

\subsection{Longer Inputs: the MRCR 256k Bin}
\label{app:mrcr256}

To test whether the advantage of our method still holds on longer inputs, we separately evaluate the 256k bin of MRCR on Qwen3.5-4B (Table~\ref{tab:mrcr256}).

\begin{table}[!ht]
  \centering
  \small
  \setlength{\tabcolsep}{6pt}
  \setlength{\aboverulesep}{0pt}
  \setlength{\belowrulesep}{0pt}
  \renewcommand{\arraystretch}{1.12}
  \begin{tabular}{l!{\color{black!40}\vrule}c!{\color{black!40}\vrule}ccc}
    \toprule
    \textbf{Training data} & \textbf{Overall} & \textbf{2-needle} & \textbf{4-needle} & \textbf{8-needle} \\
    \midrule
    LongAlign & 18.87\std{5.42} & 26.56\std{5.41} & 18.62\std{6.85} & 11.57\std{4.02} \\
    LongAlpaca & 8.21\std{3.48} & 11.18\std{4.95} & 6.25\std{4.46} & 7.28\std{1.43} \\
    LongReward & 17.11\std{1.38} & 23.79\std{3.11} & 15.55\std{1.43} & 12.14\std{0.10} \\
    SEALONG & 0.26\std{0.28} & 0.29\std{0.31} & 0.36\std{0.38} & 0.12\std{0.15} \\
    LongMIT & 15.54\std{3.35} & 21.97\std{5.14} & 13.32\std{3.50} & 11.46\std{1.63} \\
    LongMagpie & 15.86\std{1.38} & 20.98\std{1.81} & 13.89\std{1.70} & 12.81\std{1.31} \\
    LongFaith & 8.12\std{3.27} & 12.38\std{4.70} & 7.16\std{2.92} & 4.89\std{2.35} \\
    \midrule
    \rowcolor{oursbg}\textbf{\lfone{}} & \textbf{22.42}\std{1.04} & \textbf{32.00}\std{2.02} & \textbf{20.61}\std{1.12} & \textbf{14.84}\std{2.23} \\
    \bottomrule
  \end{tabular}
  \caption{Results on the 256k bin of MRCR (Qwen3.5-4B).}
  \label{tab:mrcr256}
\end{table}

\textbf{\lfone{} still leads all baselines on the 256k bin.} Its overall score is 22.42, 3.55 points above the strongest baseline, LongAlign (a relative improvement of 18.8\%), and it also ranks first on the 2-, 4-, and 8-needle groups. This is consistent with the findings of Section~\ref{sec:measurement}: the evidence usage of existing data barely grows as documents get longer, whereas our data requires answers to draw evidence from multiple places in the document, an ability that matters more as inputs get longer.

\subsection{Short-Context Ability}
\label{app:short}

Long-context fine-tuning may hurt a model's ability on short inputs. This section tests whether training on our data hurts short-context ability more than training on the baseline data.

\paragraph{Setup.}
We evaluate on nine short-context benchmarks: HellaSwag~\citep{zellers2019hellaswag}, ARC-Challenge and ARC-Easy~\citep{clark2018arc}, PIQA~\citep{bisk2020piqa}, WinoGrande~\citep{sakaguchi2021winogrande}, LogiQA~\citep{liu2020logiqa}, LAMBADA~\citep{paperno2016lambada}, MMLU~\citep{hendrycks2021mmlu}, and GSM8K~\citep{cobbe2021gsm8k}, following the short-context evaluations of LongMagpie~\citep{gao2025longmagpie} and ProLong~\citep{gao2025prolong}. Results are given in Table~\ref{tab:short}.

\begin{table}[!ht]
  \centering
  \small
  \setlength{\aboverulesep}{0pt}
  \setlength{\belowrulesep}{0pt}
  \renewcommand{\arraystretch}{1.12}
  \resizebox{\textwidth}{!}{%
  \begin{tabular}{l!{\color{black!40}\vrule}ccccccccc!{\color{black!40}\vrule}c}
    \toprule
    \textbf{Training data} & \textbf{HellaSwag} & \textbf{ARC-C} & \textbf{ARC-E} & \textbf{PIQA} & \textbf{WinoGrande} & \textbf{LogiQA} & \textbf{LAMBADA} & \textbf{MMLU} & \textbf{GSM8K} & \textbf{Avg.} \\
    \midrule
    \multicolumn{11}{>{\columncolor{groupbg}}l}{\textit{Llama3.1-8B}} \\
    \midrule
    + LongAlign & 80.09\std{0.38} & 56.09\std{2.06} & 57.48\std{1.39} & 78.56\std{0.55} & 74.19\std{0.62} & 29.70\std{0.18} & 77.20\std{0.99} & 47.10\std{4.50} & 38.62\std{2.12} & 59.89\std{0.42} \\
    + LongAlpaca & 81.79\std{0.70} & 58.76\std{1.11} & 69.73\std{0.80} & 80.38\std{0.23} & 74.88\std{0.16} & 30.67\std{2.00} & 76.28\std{0.82} & 42.69\std{6.33} & 42.30\std{1.59} & 61.94\std{0.58} \\
    + LongReward & 78.98\std{0.48} & 56.97\std{0.43} & 60.59\std{0.97} & 78.53\std{0.33} & 74.38\std{1.23} & 30.01\std{0.47} & 77.32\std{1.22} & 44.63\std{2.87} & 37.20\std{5.08} & 59.84\std{0.82} \\
    + SEALONG & 76.59\std{0.43} & 52.82\std{1.27} & 51.73\std{0.34} & 77.98\std{0.45} & 74.69\std{0.36} & 30.01\std{0.64} & 75.28\std{0.59} & 34.27\std{5.60} & 31.31\std{0.95} & 56.07\std{0.58} \\
    + LongMIT & 76.11\std{0.82} & 55.18\std{1.45} & 67.44\std{3.02} & 79.72\std{0.17} & 71.32\std{2.73} & 30.57\std{1.26} & 76.08\std{0.67} & 51.28\std{2.81} & 21.10\std{3.22} & 58.76\std{0.81} \\
    + LongMagpie & 79.96\std{0.73} & 58.25\std{0.25} & 62.89\std{1.73} & 79.51\std{0.42} & 72.61\std{1.09} & 31.08\std{1.34} & 71.32\std{1.22} & 48.47\std{9.12} & 48.09\std{2.87} & 61.35\std{0.84} \\
    + LongFaith & 78.72\std{0.35} & 56.80\std{0.55} & 56.47\std{1.90} & 78.24\std{0.38} & 75.45\std{2.05} & 29.49\std{0.55} & 76.08\std{0.56} & 43.12\std{8.06} & 39.27\std{2.96} & 59.29\std{1.31} \\
    \midrule
    \rowcolor{oursbg}+ \textbf{\lfone{}} & 79.71\std{0.16} & 57.42\std{0.48} & 69.56\std{1.69} & 79.45\std{0.17} & 72.22\std{0.75} & 29.34\std{0.86} & 77.30\std{0.50} & 53.61\std{4.62} & 43.21\std{1.74} & 62.42\std{0.50} \\
    \rowcolor{oursbg}+ \textbf{\lfn{}} & 79.45\std{0.38} & 58.39\std{0.44} & 72.39\std{1.91} & 79.92\std{0.76} & 71.67\std{0.82} & 30.52\std{0.79} & 79.02\std{0.60} & 60.06\std{1.17} & 48.72\std{0.57} & 64.46\std{0.15} \\
    \midrule
    \multicolumn{11}{>{\columncolor{groupbg}}l}{\textit{Qwen3.5-4B}} \\
    \midrule
    + LongAlign & 73.20\std{0.98} & 61.92\std{1.15} & 63.87\std{1.39} & 75.79\std{0.58} & 70.88\std{1.99} & 32.92\std{0.85} & 69.29\std{1.61} & 72.72\std{0.92} & 79.73\std{2.09} & 66.70\std{0.71} \\
    + LongAlpaca & 76.47\std{0.32} & 62.54\std{0.48} & 70.85\std{2.18} & 77.04\std{0.11} & 71.43\std{0.34} & 34.46\std{0.47} & 68.02\std{1.47} & 74.45\std{0.12} & 83.55\std{0.66} & 68.76\std{0.20} \\
    + LongReward & 73.03\std{0.14} & 60.58\std{0.48} & 66.74\std{0.52} & 76.53\std{0.13} & 69.30\std{0.65} & 33.03\std{0.41} & 68.50\std{1.67} & 72.16\std{0.11} & 74.98\std{7.03} & 66.09\std{0.99} \\
    + SEALONG & 70.27\std{0.74} & 53.19\std{3.11} & 52.67\std{3.94} & 74.30\std{0.64} & 69.38\std{0.75} & 31.59\std{0.69} & 68.56\std{0.50} & 68.56\std{2.60} & 53.35\std{5.78} & 60.21\std{1.87} \\
    + LongMIT & 71.61\std{1.04} & 58.82\std{1.35} & 68.50\std{1.47} & 76.84\std{0.58} & 69.30\std{0.96} & 34.46\std{0.69} & 69.03\std{1.09} & 69.68\std{0.74} & 71.95\std{2.40} & 65.58\std{0.06} \\
    + LongMagpie & 76.50\std{0.29} & 63.05\std{0.74} & 65.73\std{1.81} & 77.51\std{0.30} & 72.45\std{0.69} & 33.08\std{0.49} & 68.39\std{0.98} & 75.44\std{0.29} & 80.62\std{0.59} & 68.08\std{0.23} \\
    + LongFaith & 71.50\std{0.42} & 57.45\std{1.92} & 65.00\std{1.96} & 76.24\std{0.17} & 71.11\std{0.00} & 31.90\std{0.58} & 70.89\std{0.32} & 71.30\std{2.99} & 67.80\std{1.48} & 64.80\std{0.68} \\
    \midrule
    \rowcolor{oursbg}+ \textbf{\lfone{}} & 71.88\std{1.16} & 64.79\std{1.11} & 73.88\std{1.91} & 75.44\std{0.23} & 70.27\std{0.73} & 32.41\std{0.96} & 68.53\std{0.35} & 73.97\std{0.49} & 80.59\std{0.95} & 67.97\std{0.30} \\
    \rowcolor{oursbg}+ \textbf{\lfn{}} & 70.60\std{0.38} & 65.33\std{0.49} & 76.12\std{0.99} & 75.63\std{0.43} & 70.64\std{0.89} & 33.38\std{0.58} & 68.78\std{0.87} & 75.21\std{0.12} & 79.53\std{0.67} & 68.36\std{0.07} \\
    \bottomrule
  \end{tabular}}
  \caption{Results on nine short-context benchmarks.}
  \label{tab:short}
\end{table}

\paragraph{Results.}
Our two mixtures do not hurt short-context ability more than the baselines do. On Llama3.1-8B, the averages of \lfone{} and \lfn{} are 62.42 and 64.46, higher than all seven baselines; on Qwen3.5-4B, they are 67.97 and 68.36, within 0.8 points of the strongest baseline, LongAlpaca (68.76), and higher than most other baselines.

\subsection{Effect of Training Size}
\label{app:scale}

All experiments in the main text fix 2{,}048 training samples. This section tests whether enlarging the training set brings further gains, with the same setting as the ablations in Section~\ref{sec:ablation} (\lfone{}, Qwen3.5-4B). The training data takes 2{,}048, 4{,}096, and 8{,}192 samples in turn, each size containing all samples of the previous one; results are given in Table~\ref{tab:scale}.

\begin{table}[!ht]
  \centering
  \small
  \setlength{\tabcolsep}{4pt}
  \setlength{\aboverulesep}{0pt}
  \setlength{\belowrulesep}{0pt}
  \renewcommand{\arraystretch}{1.12}
  \begin{tabular}{l!{\color{black!40}\vrule}cccccc!{\color{black!40}\vrule}c}
    \toprule
    \textbf{Training samples} & \textbf{LongBench} & \textbf{LongBench v2} & \textbf{LooGLE} & \textbf{MRCR} & \textbf{HELMET} & \textbf{RULER} & \textbf{Avg.} \\
    \midrule
    \rowcolor{groupbg}2048 (control) & 46.34\std{0.35} & \textbf{37.00}\std{0.20} & \textbf{71.39}\std{0.56} & 35.18\std{1.25} & \textbf{60.61}\std{0.61} & 90.87\std{0.56} & 56.90\std{0.10} \\
    \rowcolor{oursbg}4096 & 47.94\std{1.21} & 36.73\std{1.81} & 71.27\std{0.83} & \textbf{36.07}\std{1.53} & 60.30\std{0.86} & \textbf{91.20}\std{0.83} & \textbf{57.25}\std{0.33} \\
    \rowcolor{oursbg}8192 & \textbf{48.55}\std{0.36} & 36.80\std{0.92} & 70.00\std{0.78} & 35.50\std{0.74} & 60.19\std{0.40} & 91.08\std{0.93} & 57.02\std{0.37} \\
    \bottomrule
  \end{tabular}
  \caption{Results on six benchmarks as the training size grows from 2{,}048 to 8{,}192 samples (\lfone{}, Qwen3.5-4B).}
  \label{tab:scale}
\end{table}

\textbf{Scaling up brings no clear gain.} Quadrupling the data changes the six-benchmark average by only $+0.12$ ($+0.35$ and $-0.23$ between adjacent sizes). Individual benchmarks show no consistent trend either: LongBench rises with data size (46.34 to 48.55) while LooGLE falls (71.39 to 70.00), and the two cancel out in the average. This is consistent with the findings of LIMA~\citep{zhou2023lima} and related work: a small amount of high-quality instruction data achieves near-optimal results, and adding more data from the same distribution brings limited benefit.

This result supports fixing 2{,}048 samples in the main text: the comparisons in the main table are not affected by the choice of training size. In contrast, \lfn{}, also with 2{,}048 samples, only organizes multiple QA pairs on the same document into a multi-turn dialogue, yet its six-benchmark average is 1.76 and 0.97 points higher than that of \lfone{}, exceeding the change brought by quadrupling the data. At the scales we test, the key to improving long-context data thus lies in the structure of the data rather than its quantity.

\FloatBarrier
\section{Per-Benchmark Breakdown of the Main Results}
\label{app:breakdown}

Tables~\ref{tab:loogle}--\ref{tab:helmet} give the subtask-level results behind Table~\ref{tab:main}, for LooGLE, LongBench, LongBench v2, MRCR, and HELMET in turn.

\begin{table}[!ht]
  \centering
  \small
  \setlength{\aboverulesep}{0pt}
  \setlength{\belowrulesep}{0pt}
  \renewcommand{\arraystretch}{1.12}
  \begin{tabular}{l!{\color{black!40}\vrule}cc}
    \toprule
    \textbf{Training data} & \textbf{Short-dep QA} & \textbf{Long-dep QA} \\
    \midrule
    \multicolumn{3}{>{\columncolor{groupbg}}l}{\textit{Llama3.1-8B}} \\
    \midrule
    + LongAlign & 82.62\std{0.98} & 40.72\std{1.36} \\
    + LongAlpaca & 83.13\std{0.18} & 41.03\std{1.29} \\
    + LongReward & 80.94\std{0.22} & 42.63\std{0.41} \\
    + SEALONG & 59.45\std{3.83} & 27.25\std{2.54} \\
    + LongMIT & 84.67\std{1.48} & 41.35\std{2.57} \\
    + LongMagpie & 78.90\std{3.16} & 42.54\std{1.22} \\
    + LongFaith & 73.01\std{0.43} & 30.46\std{1.05} \\
    \midrule
    \rowcolor{oursbg}+ \textbf{\lfone{}} & 85.56\std{0.53} & 44.35\std{1.05} \\
    \rowcolor{oursbg}+ \textbf{\lfn{}} & \textbf{85.90}\std{0.47} & \textbf{47.26}\std{0.61} \\
    \midrule
    \multicolumn{3}{>{\columncolor{groupbg}}l}{\textit{Qwen3.5-4B}} \\
    \midrule
    + LongAlign & 82.54\std{1.20} & 49.11\std{1.95} \\
    + LongAlpaca & 85.48\std{0.63} & 53.10\std{1.44} \\
    + LongReward & 83.07\std{1.20} & 49.83\std{0.99} \\
    + SEALONG & 62.91\std{1.73} & 36.54\std{0.82} \\
    + LongMIT & 84.74\std{0.25} & 45.90\std{0.55} \\
    + LongMagpie & 80.73\std{0.41} & 51.07\std{1.37} \\
    + LongFaith & 82.45\std{0.70} & 47.44\std{1.00} \\
    \midrule
    \rowcolor{oursbg}+ \textbf{\lfone{}} & \textbf{88.55}\std{0.36} & 54.22\std{1.03} \\
    \rowcolor{oursbg}+ \textbf{\lfn{}} & 88.07\std{0.39} & \textbf{54.49}\std{1.64} \\
    \bottomrule
  \end{tabular}
  \caption{Results on the two QA subtasks of LooGLE.}
  \label{tab:loogle}
\end{table}

\begin{table}[!ht]
  \centering
  \small
  \setlength{\aboverulesep}{0pt}
  \setlength{\belowrulesep}{0pt}
  \renewcommand{\arraystretch}{1.12}
  \setlength{\tabcolsep}{4pt}
  \resizebox{\textwidth}{!}{%
  \begin{tabular}{l!{\color{black!40}\vrule}c!{\color{black!40}\vrule}cccccc}
    \toprule
    \textbf{Training data} & \textbf{Avg} & \textbf{Single-Doc QA} & \textbf{Multi-Doc QA} & \textbf{Summarization} & \textbf{Few-shot} & \textbf{Synthetic} & \textbf{Code} \\
    \midrule
    \multicolumn{8}{>{\columncolor{groupbg}}l}{\textit{Llama3.1-8B}} \\
    \midrule
    + LongAlign & 35.37\std{0.53} & 17.44\std{0.12} & 13.95\std{0.43} & \textbf{24.75}\std{0.20} & 60.45\std{1.31} & 50.13\std{2.51} & 63.05\std{2.76} \\
    + LongAlpaca & 30.78\std{0.27} & 16.74\std{0.44} & 13.54\std{0.89} & 21.12\std{1.38} & 61.90\std{0.34} & 20.54\std{6.31} & 65.80\std{2.33} \\
    + LongReward & 35.28\std{0.73} & 19.52\std{0.19} & 14.46\std{0.16} & 24.60\std{0.16} & 62.63\std{0.95} & 39.80\std{5.42} & 68.26\std{2.13} \\
    + SEALONG & 23.02\std{0.61} & 8.17\std{0.64} & 3.31\std{0.17} & 17.16\std{0.59} & 60.66\std{0.92} & 2.42\std{1.64} & 59.48\std{7.89} \\
    + LongMIT & 33.69\std{0.91} & \textbf{32.78}\std{2.67} & \textbf{20.04}\std{4.90} & 22.76\std{0.29} & 60.35\std{0.46} & 12.07\std{7.13} & 63.84\std{1.08} \\
    + LongMagpie & 35.92\std{0.20} & 18.38\std{0.90} & 13.93\std{0.56} & 23.79\std{0.61} & 61.59\std{0.68} & 47.12\std{4.55} & \textbf{71.08}\std{1.34} \\
    + LongFaith & 23.67\std{0.42} & 8.49\std{0.49} & 3.25\std{0.15} & 18.83\std{0.02} & 62.12\std{1.61} & 1.17\std{1.00} & 61.36\std{2.37} \\
    \midrule
    \rowcolor{oursbg}+ \textbf{\lfone{}} & 41.56\std{0.09} & 28.36\std{1.56} & 18.90\std{0.88} & 24.18\std{0.20} & \textbf{63.29}\std{0.36} & \textbf{65.76}\std{0.38} & 68.28\std{1.19} \\
    \rowcolor{oursbg}+ \textbf{\lfn{}} & \textbf{42.27}\std{1.10} & 32.27\std{1.11} & 19.81\std{1.72} & 24.29\std{0.09} & 62.96\std{0.98} & 64.93\std{1.92} & 67.81\std{1.29} \\
    \midrule
    \multicolumn{8}{>{\columncolor{groupbg}}l}{\textit{Qwen3.5-4B}} \\
    \midrule
    + LongAlign & 35.74\std{0.33} & 17.60\std{0.22} & 13.47\std{0.55} & \textbf{25.52}\std{0.27} & 60.93\std{0.88} & 50.42\std{2.62} & 64.63\std{1.83} \\
    + LongAlpaca & 40.61\std{1.30} & 28.92\std{3.14} & 18.05\std{3.26} & 21.04\std{0.51} & 62.92\std{0.48} & 65.84\std{1.59} & 65.88\std{2.09} \\
    + LongReward & 38.62\std{1.73} & 25.83\std{2.98} & 16.52\std{1.74} & 25.41\std{0.47} & 62.61\std{0.54} & 54.38\std{8.18} & 63.19\std{2.33} \\
    + SEALONG & 26.50\std{0.21} & 12.67\std{0.91} & 8.35\std{0.57} & 22.60\std{0.16} & 61.89\std{0.41} & 0.28\std{0.25} & 66.86\std{0.29} \\
    + LongMIT & 40.58\std{1.03} & 42.30\std{1.28} & 23.92\std{1.67} & 23.08\std{0.08} & 59.34\std{0.97} & 41.89\std{7.08} & 66.02\std{4.03} \\
    + LongMagpie & 45.42\std{0.10} & 36.52\std{0.80} & 25.38\std{0.83} & 24.96\std{0.02} & \textbf{64.19}\std{0.18} & 69.06\std{0.84} & \textbf{71.17}\std{0.37} \\
    + LongFaith & 28.75\std{0.26} & 19.46\std{1.13} & 9.57\std{0.13} & 23.78\std{0.20} & 62.80\std{0.31} & 0.43\std{0.34} & 70.00\std{0.35} \\
    \midrule
    \rowcolor{oursbg}+ \textbf{\lfone{}} & 46.34\std{0.35} & 40.93\std{1.76} & 39.09\std{0.93} & 23.91\std{0.13} & 61.24\std{1.03} & 70.44\std{1.17} & 50.54\std{2.65} \\
    \rowcolor{oursbg}+ \textbf{\lfn{}} & \textbf{47.69}\std{0.19} & \textbf{43.03}\std{0.14} & \textbf{43.01}\std{0.66} & 24.09\std{0.08} & 60.55\std{0.51} & \textbf{71.83}\std{1.61} & 51.64\std{2.13} \\
    \bottomrule
  \end{tabular}}
  \caption{Subtask results on LongBench.}
  \label{tab:longbench}
\end{table}

\begin{table}[!ht]
  \centering
  \small
  \setlength{\aboverulesep}{0pt}
  \setlength{\belowrulesep}{0pt}
  \renewcommand{\arraystretch}{1.12}
  \begin{tabular}{l!{\color{black!40}\vrule}c!{\color{black!40}\vrule}ccccc}
    \toprule
    \textbf{Training data} & \textbf{Overall} & \textbf{Easy} & \textbf{Hard} & \textbf{Short} & \textbf{Medium} & \textbf{Long} \\
    \midrule
    \multicolumn{7}{>{\columncolor{groupbg}}l}{\textit{Llama3.1-8B}} \\
    \midrule
    + LongAlign & 25.23\std{1.66} & 25.17\std{2.12} & 25.30\std{1.45} & 27.40\std{3.59} & 24.07\std{0.71} & 24.10\std{2.80} \\
    + LongAlpaca & 26.17\std{3.92} & 26.73\std{5.62} & 25.80\std{3.05} & 28.87\std{4.72} & 23.87\std{3.71} & 26.23\std{3.52} \\
    + LongReward & 26.53\std{0.70} & 27.60\std{0.87} & 25.93\std{1.19} & 28.17\std{3.07} & 25.73\std{0.71} & 25.63\std{1.93} \\
    + SEALONG & 19.23\std{1.45} & 20.33\std{1.55} & 18.53\std{2.56} & 18.33\std{2.39} & 19.23\std{2.41} & 20.70\std{6.17} \\
    + LongMIT & 24.27\std{0.67} & 23.97\std{1.55} & 24.43\std{2.12} & 21.30\std{3.73} & 25.90\std{1.49} & 25.93\std{6.11} \\
    + LongMagpie & 29.87\std{2.04} & 32.63\std{5.08} & 28.20\std{0.46} & 35.53\std{2.97} & \textbf{27.90}\std{2.01} & 24.40\std{3.50} \\
    + LongFaith & 20.77\std{3.31} & 24.80\std{1.66} & 18.20\std{4.35} & 20.37\std{1.96} & 20.30\std{4.66} & 22.23\std{4.28} \\
    \midrule
    \rowcolor{oursbg}+ \textbf{\lfone{}} & 28.20\std{1.78} & 29.70\std{2.10} & 27.33\std{1.67} & 32.37\std{2.63} & 26.50\std{0.50} & 24.70\std{3.74} \\
    \rowcolor{oursbg}+ \textbf{\lfn{}} & \textbf{33.00}\std{1.20} & \textbf{37.30}\std{1.31} & \textbf{30.33}\std{1.11} & \textbf{38.73}\std{2.73} & 27.43\std{0.81} & \textbf{34.57}\std{4.73} \\
    \midrule
    \multicolumn{7}{>{\columncolor{groupbg}}l}{\textit{Qwen3.5-4B}} \\
    \midrule
    + LongAlign & 30.27\std{0.12} & 32.60\std{1.21} & 28.83\std{0.92} & 36.10\std{1.70} & 27.73\std{1.43} & 25.60\std{3.26} \\
    + LongAlpaca & 33.80\std{1.51} & 34.57\std{1.70} & 33.33\std{1.85} & 37.03\std{1.14} & 32.57\std{1.21} & 30.90\std{3.26} \\
    + LongReward & 33.20\std{1.00} & 33.70\std{2.48} & 32.90\std{1.93} & 37.43\std{1.27} & 31.17\std{0.95} & 30.23\std{5.27} \\
    + SEALONG & 28.27\std{0.23} & 29.13\std{2.76} & 27.77\std{1.36} & 27.03\std{1.14} & 30.67\std{2.14} & 25.60\std{2.36} \\
    + LongMIT & 30.27\std{0.31} & 30.57\std{1.10} & 30.10\std{0.17} & 30.57\std{1.47} & 30.37\std{0.29} & 29.63\std{1.85} \\
    + LongMagpie & 34.33\std{1.33} & 33.17\std{2.68} & 35.03\std{0.95} & 39.80\std{1.15} & 33.07\std{0.81} & 27.80\std{2.80} \\
    + LongFaith & 36.07\std{1.53} & 37.70\std{2.08} & 35.03\std{1.70} & 37.20\std{1.10} & \textbf{36.27}\std{2.03} & 33.63\std{3.21} \\
    \midrule
    \rowcolor{oursbg}+ \textbf{\lfone{}} & 37.00\std{0.20} & 36.97\std{1.55} & 37.00\std{1.18} & 42.60\std{1.41} & 34.27\std{0.23} & 33.03\std{2.66} \\
    \rowcolor{oursbg}+ \textbf{\lfn{}} & \textbf{38.93}\std{0.81} & \textbf{39.77}\std{2.84} & \textbf{38.37}\std{0.50} & \textbf{44.63}\std{1.68} & 35.67\std{1.10} & \textbf{35.80}\std{1.37} \\
    \bottomrule
  \end{tabular}
  \caption{Subtask results on LongBench v2, sliced by difficulty and length.}
  \label{tab:longbenchv2}
\end{table}

\begin{table}[!ht]
  \centering
  \small
  \setlength{\aboverulesep}{0pt}
  \setlength{\belowrulesep}{0pt}
  \renewcommand{\arraystretch}{1.12}
  \begin{tabular}{l!{\color{black!40}\vrule}c!{\color{black!40}\vrule}ccccc}
    \toprule
    \textbf{Training data} & \textbf{Avg} & \textbf{8K} & \textbf{16K} & \textbf{32K} & \textbf{64K} & \textbf{128K}  \\
    \midrule
    \multicolumn{7}{>{\columncolor{groupbg}}l}{\textit{Llama3.1-8B}} \\
    \midrule
    + LongAlign & 1.54\std{1.29} & 3.65\std{2.94} & 1.64\std{1.60} & 0.82\std{0.74} & 1.29\std{1.01} & 0.13\std{0.20}  \\
    + LongAlpaca & 0.02\std{0.03} & 0.00\std{0.00} & 0.00\std{0.00} & 0.00\std{0.00} & 0.13\std{0.18} & 0.00\std{0.00}  \\
    + LongReward & 4.60\std{2.57} & 2.64\std{1.76} & 6.06\std{3.93} & 5.73\std{3.71} & 8.23\std{3.88} & 0.58\std{0.56}  \\
    + SEALONG & 0.85\std{0.73} & 0.94\std{0.78} & 1.45\std{1.24} & 1.39\std{1.62} & 0.47\std{0.41} & 0.00\std{0.00}  \\
    + LongMIT & 11.98\std{3.15} & 17.52\std{5.73} & 18.58\std{3.34} & 12.66\std{4.94} & 8.67\std{1.31} & 2.12\std{0.36}  \\
    + LongMagpie & 12.55\std{6.89} & \textbf{18.06}\std{1.59} & 15.04\std{9.02} & 16.59\std{8.82} & 8.32\std{9.42} & 4.32\std{6.97}  \\
    + LongFaith & 0.00\std{0.00} & 0.00\std{0.00} & 0.00\std{0.00} & 0.00\std{0.00} & 0.00\std{0.00} & 0.00\std{0.00}  \\
    \midrule
    \rowcolor{oursbg}+ \textbf{\lfone{}} & \textbf{19.29}\std{2.18} & 16.52\std{2.69} & \textbf{27.18}\std{6.08} & \textbf{24.92}\std{2.92} & \textbf{21.20}\std{0.61} & \textbf{7.10}\std{1.54}  \\
    \rowcolor{oursbg}+ \textbf{\lfn{}} & 15.92\std{1.84} & 12.96\std{3.42} & 23.12\std{1.82} & 20.48\std{2.88} & 19.58\std{2.38} & 3.93\std{0.22}  \\
    \midrule
    \multicolumn{7}{>{\columncolor{groupbg}}l}{\textit{Qwen3.5-4B}} \\
    \midrule
    + LongAlign & 30.07\std{3.48} & 34.51\std{1.35} & 32.54\std{2.94} & 29.84\std{3.06} & 29.34\std{6.25} & 23.80\std{4.08}  \\
    + LongAlpaca & 18.90\std{3.63} & 19.33\std{3.24} & 20.99\std{4.39} & 19.61\std{5.60} & 21.39\std{4.29} & 13.23\std{1.61}  \\
    + LongReward & 32.38\std{0.77} & 34.06\std{1.67} & 34.23\std{2.37} & 36.03\std{0.70} & 34.08\std{0.83} & 23.49\std{1.08}  \\
    + SEALONG & 2.48\std{2.89} & 3.66\std{5.02} & 2.30\std{3.29} & 3.37\std{3.56} & 1.83\std{1.71} & 1.12\std{1.02}  \\
    + LongMIT & 30.32\std{0.83} & 34.19\std{1.02} & 32.38\std{1.49} & 32.94\std{1.02} & 28.43\std{1.89} & 23.37\std{2.54}  \\
    + LongMagpie & 32.75\std{0.99} & 35.50\std{0.67} & 35.00\std{2.02} & 34.52\std{1.35} & 34.94\std{1.39} & 23.67\std{0.90}  \\
    + LongFaith & 22.00\std{5.24} & 29.83\std{4.29} & 24.54\std{3.35} & 21.37\std{6.53} & 17.57\std{6.77} & 16.06\std{5.63}  \\
    \midrule
    \rowcolor{oursbg}+ \textbf{\lfone{}} & 35.18\std{1.25} & 39.09\std{3.22} & 38.49\std{1.44} & 36.72\std{1.09} & 33.51\std{2.52} & \textbf{27.81}\std{1.79}  \\
    \rowcolor{oursbg}+ \textbf{\lfn{}} & \textbf{37.08}\std{0.43} & \textbf{42.81}\std{0.25} & \textbf{41.01}\std{1.86} & \textbf{37.50}\std{0.02} & \textbf{36.18}\std{0.92} & 27.54\std{1.45}  \\
    \bottomrule
  \end{tabular}
  \caption{Per-bin results on MRCR up to 128k. Avg.\ is pooled over all samples in these five bins; the 256k bin is evaluated separately in Appendix~\ref{app:mrcr256}.}
  \label{tab:mrcr}
\end{table}

\begin{table}[!ht]
  \centering
  \small
  \setlength{\aboverulesep}{0pt}
  \setlength{\belowrulesep}{0pt}
  \renewcommand{\arraystretch}{1.12}
  \begin{tabular}{l!{\color{black!40}\vrule}c!{\color{black!40}\vrule}cccc}
    \toprule
    \textbf{Training data} & \textbf{Avg} & \textbf{RAG} & \textbf{ICL} & \textbf{Re-rank} & \textbf{LongQA} \\
    \midrule
    \multicolumn{6}{>{\columncolor{groupbg}}l}{\textit{Llama3.1-8B}} \\
    \midrule
    + LongAlign & 42.15\std{1.39} & 49.06\std{3.03} & 78.91\std{3.34} & 29.55\std{4.83} & 11.05\std{3.04} \\
    + LongAlpaca & 42.91\std{1.47} & 59.64\std{0.38} & 72.10\std{3.50} & 26.14\std{1.39} & 13.75\std{1.08} \\
    + LongReward & 49.07\std{1.51} & 59.34\std{1.40} & 80.36\std{2.16} & 31.57\std{1.93} & 25.01\std{4.40} \\
    + SEALONG & 44.32\std{3.51} & 49.48\std{4.71} & 76.59\std{0.63} & 30.00\std{4.98} & 21.19\std{3.86} \\
    + LongMIT & 46.71\std{0.93} & 60.16\std{0.27} & 77.75\std{1.24} & 18.02\std{2.28} & 30.90\std{0.79} \\
    + LongMagpie & 50.70\std{2.00} & 60.24\std{0.34} & 78.33\std{4.06} & 32.26\std{3.26} & \textbf{31.96}\std{1.91} \\
    + LongFaith & 47.05\std{3.54} & 55.77\std{7.78} & 76.12\std{2.26} & 34.73\std{1.20} & 21.56\std{3.81} \\
    \midrule
    \rowcolor{oursbg}+ \textbf{\lfone{}} & 53.35\std{0.85} & 64.31\std{0.62} & 81.53\std{2.99} & 39.06\std{4.20} & 28.50\std{5.11} \\
    \rowcolor{oursbg}+ \textbf{\lfn{}} & \textbf{56.71}\std{0.39} & \textbf{64.89}\std{0.37} & \textbf{84.00}\std{0.29} & \textbf{47.16}\std{0.75} & 30.80\std{1.51} \\
    \midrule
    \multicolumn{6}{>{\columncolor{groupbg}}l}{\textit{Qwen3.5-4B}} \\
    \midrule
    + LongAlign & 51.15\std{2.73} & 52.81\std{6.75} & 83.96\std{1.01} & 43.03\std{0.94} & 24.79\std{3.08} \\
    + LongAlpaca & 57.11\std{0.37} & 64.12\std{1.06} & 80.89\std{0.86} & 45.95\std{1.11} & 37.47\std{1.00} \\
    + LongReward & 55.59\std{0.21} & 65.02\std{0.35} & 77.67\std{2.45} & 40.47\std{3.26} & 39.20\std{0.35} \\
    + SEALONG & 52.38\std{1.27} & 53.50\std{2.24} & 82.97\std{0.51} & 40.77\std{0.53} & 32.29\std{2.26} \\
    + LongMIT & 55.23\std{0.87} & 65.24\std{0.47} & 83.17\std{0.61} & 36.53\std{2.34} & 35.99\std{1.03} \\
    + LongMagpie & 57.63\std{0.87} & 65.82\std{0.29} & 83.11\std{0.71} & 40.28\std{2.34} & 41.31\std{0.49} \\
    + LongFaith & 60.19\std{0.23} & 65.54\std{0.07} & \textbf{84.14}\std{0.26} & 50.55\std{0.89} & 40.53\std{0.20} \\
    \midrule
    \rowcolor{oursbg}+ \textbf{\lfone{}} & 60.61\std{0.61} & 66.79\std{0.44} & 83.60\std{0.75} & 51.16\std{1.04} & 40.88\std{0.68} \\
    \rowcolor{oursbg}+ \textbf{\lfn{}} & \textbf{60.95}\std{0.88} & \textbf{67.55}\std{0.30} & 79.51\std{1.50} & \textbf{52.40}\std{2.00} & \textbf{44.31}\std{0.38} \\
    \bottomrule
  \end{tabular}
  \caption{Results on the four HELMET categories.}
  \label{tab:helmet}
\end{table}

\end{document}